\documentclass[sn-mathphys,Numbered]{sn-jnl}

\usepackage{graphicx}
\usepackage{booktabs}
\usepackage{longtable}
\usepackage{array}
\usepackage{amsmath,amssymb}
\usepackage{siunitx}
\usepackage{microtype}
\usepackage{manyfoot}
\usepackage{placeins}
\usepackage{xcolor}
\usepackage{tabularx}

\hypersetup{hidelinks}
\newcommand{\keyresult}[1]{%
\begin{center}
\setlength{\fboxsep}{8pt}%
\fcolorbox{black}{gray!5}{%
\begin{minipage}{0.92\linewidth}
\textbf{Key finding.} #1
\end{minipage}}
\end{center}}

\begin{document}

\title{Adaptive Rotation for iSOMA: Geometry, Benchmarking, and Noise Robustness in Variational Quantum Objectives}

\author*[1,2,3]{\fnm{Vojtěch} \sur{Novák}}\email{vojtech.novak.st1@vsb.cz}

\author[1,2,3]{\fnm{Ivan} \sur{Zelinka}}

\affil*[1]{\orgdiv{Department of Computer Science, Faculty of Electrical Engineering and Computer Science}, \orgname{VSB-Technical University of Ostrava}, \city{Ostrava}, \country{Czech Republic}}

\affil[2]{\orgname{IT4Innovations National Supercomputing Center}, \orgname{VSB-Technical University of Ostrava}, \city{Ostrava}, \postcode{708 00}, \country{Czech Republic}}

\affil[3]{\orgdiv{Department of Informatics and Statistics, Marine Research Institute}, \orgname{Klaipeda University}, \city{Klaipeda}, \country{Lithuania}}

\abstract{We study whether the coordinate dependence of the improved Self-Organizing Migrating Algorithm (iSOMA) can be reduced while retaining its inexpensive leader-directed migration mechanism. We introduce iSOMA-AR, which learns a basis from successful migration displacements and selectively applies the standard perturbation mask in that basis. On the complete noiseless BBOB suite, iSOMA-AR significantly outperformed baseline iSOMA across matched conditions, with the largest gains on geometrically difficult landscapes. A targeted ablation shows that the learned orientation is beneficial on a rotated ill-conditioned landscape and that moderate changes of the gate threshold and rotation cap preserve the qualitative result. On CEC 2011 Real
World Optimization Problems, iSOMA-AR outperformed iL-SHADE on most problems, although its advantage over baseline iSOMA was not statistically significant. A canonical-jSO rerun is reported as a post-hoc sensitivity check alongside the original jSO-derived comparator. On frustrated-spin variational quantum objectives, adaptive rotation improved most transverse-field conditions, while gains on the diagonal and anisotropic models were absent or selective. Under strong effective sampling noise, the SOMA variants were the most robust population-based methods in the comparison, but iSOMA-AR was not significantly better than baseline iSOMA. Repairing all-zero PRT masks greatly reduced repeated-point evaluations without changing endpoint quality significantly, making this implementation detail unlikely to explain the noise result. Overall, adaptive rotation is most useful on coordinate-sensitive deterministic problems, while the observed noise robustness appears to arise mainly from the underlying SOMA migration mechanism.}

\maketitle

\section{Introduction}

Continuous black-box global optimization becomes difficult not only as dimensionality grows, but also when the objective departs from an isotropic, separable, and deterministic landscape. Ill-conditioning, interactions between variables, rotated valleys and ridges, multimodality, weak global structure, ruggedness, and noisy function values can all change which search directions are informative and how reliably candidate solutions can be ranked. These issues motivate population-based derivative-free methods, which trade local derivative information for broader exploration and are widely used when gradients are unavailable, unreliable, or too expensive to obtain. Differential Evolution (DE) is a canonical example of this class \citep{storn1997de,das2011desurvey,das2016deupdated}, while covariance-adapting evolution strategies illustrate the value of explicitly learning dependencies between variables on badly scaled and non-separable problems \citep{hansen2001cma}. The resulting challenge is not simply to search globally, but to do so with proposal mechanisms that remain effective across substantially different landscape geometries.

The Self-Organizing Migrating Algorithm (SOMA) approaches this problem through leader-directed migration rather than the mutation--crossover generations used by DE. A migrant evaluates points along a path toward a selected leader, while a stochastic perturbation (PRT) vector activates only a subset of coordinates \citep{zelinka2004soma}. Subsequent SOMA research has focused on migration strategies, parameter adaptation, diversity preservation, and search-space control; representative developments include self-adapting SOMA and iSOMA \citep{skanderova2019sasoma,diep2022isoma}, with the broader literature reviewed in \citep{skanderova2023somareview}. These developments improve how SOMA chooses leaders, path parameters, and search regions, but the usual binary PRT mechanism remains intrinsically tied to the coordinate system: when only some coordinates are active, a migration step is restricted to coordinate-aligned subspaces. This is benign when the native coordinates reflect the problem structure, but can be inefficient when useful directions are oblique combinations of variables.

DE provides a useful parallel because its development has repeatedly exposed both the benefits and the limitations of coordinate-wise operators. A major line of work replaced fixed control parameters with online adaptation or success-history mechanisms, including self-adaptive DE, JADE, SHADE, and L-SHADE \citep{brest2006jde,zhang2009jade,tanabe2013shade,tanabe2014lshade}. Successive CEC real-parameter competitions helped consolidate this adaptive-DE lineage, from L-SHADE to iL-SHADE and jSO \citep{brest2016ilshade,brest2017jso}, and DE surveys document the broader shift toward adaptive parameter control, population-size control, archives, and hybrid search mechanisms \citep{das2011desurvey,das2016deupdated}. At the same time, DE also illustrates why adaptation of scalar parameters is not sufficient for every landscape. Differential mutation is naturally compatible with rotated coordinates, whereas conventional component-wise crossover is not; covariance- and eigenvector-based crossover operators were therefore introduced to recover performance on correlated and rotated problems \citep{guo2015eigde,caraffini2019rotation}. This literature suggests a more general design principle: when an optimizer contains a coordinate-selective operator, learning a useful basis can be as important as adapting the operator's scalar parameters.

Benchmark design makes this distinction measurable. The noiseless BBOB testbed deliberately spans separable functions, low- and high-conditioning regimes, and multimodal landscapes with both stronger and weaker global structure \citep{finck2009bbob}; the COCO methodology complements such function classes with multiple instances, target-based runtimes, and evaluation-count-based performance assessment \citep{hansen2021coco}. CEC benchmarks \citep{novakCEC2026} provide a different perspective. Beyond the artificial real-parameter suites that have driven much of modern DE development, the CEC 2011 Real-World Optimization Problems collect 22 heterogeneous application problems involving, among others, parameter estimation, molecular potentials, control, antenna design, power systems, and spacecraft trajectories \citep{das2010cec2011}. BBOB and CEC 2011 are therefore complementary here: the former isolates geometric failure modes under controlled transformations, whereas the latter tests whether an improvement transfers to application-oriented objectives with heterogeneous scales, dimensions, and structure.

Variational quantum algorithms add a further difficulty because optimization geometry and evaluation noise appear together. In the Variational Quantum Eigensolver (VQE), a classical optimizer repeatedly updates circuit parameters using energy estimates returned by a quantum evaluation loop; optimizer behavior is consequently part of the practical trainability of the method \citep{cerezo2021vqa,tilly2022vqe}. VQA landscapes can exhibit vanishing-gradient regions, and gradient-free optimization does not in general evade the associated loss of resolvable cost differences \citep{mcclean2018barren,arrasmith2021gradientfree}. Even away from asymptotic barren-plateau regimes \citep{illesovaQMetric2025}, finite-shot sampling \citep{novakNoisy2026, illesovaVHA2025, illesovaStat2025} makes objective values stochastic, so the relative performance of classical optimizers can depend strongly on the problem, sampling budget, and optimizer configuration \citep{bonetmonroig2023optimizers, novakReliable2025, bezdekOptimization2025}. Such objectives are therefore a useful stress test for population-based global search: an optimizer must make progress on a structured, potentially multimodal landscape while avoiding excessive reliance on small and noisy objective differences \citep{novakGlobalSearch}.

Against this background, the present study asks whether the coordinate dependence of an improved SOMA baseline can be reduced without replacing its inexpensive leader-directed migration mechanism by a full population model. We introduce \emph{iSOMA-AR}, which estimates a covariance matrix from successful normalized migration displacements and, when off-axis structure is sufficiently strong, applies the existing PRT mask in the learned basis. The leader selection, path construction, acceptance rule, and objective-evaluation budget remain unchanged; only the coordinate system in which masking is performed is adapted. This yields four questions: whether adaptive rotation improves iSOMA across the complete noiseless BBOB suite; which landscape classes account for gains and failures; whether the effect transfers to the CEC 2011 real-world problems; and whether the same geometry-dependent behavior persists on frustrated-spin variational quantum objectives under exact and noisy evaluations. The comparator set spans CMA-ES, iL-SHADE, a jSO-derived implementation with canonical jSO used as a sensitivity check, and SPSA where noise is present \citep{spall1992spsa}, thereby contrasting the proposed geometric correction with covariance adaptation, modern adaptive DE, and stochastic approximation.

\section{Methods}

\subsection{Canonical SOMA, iSOMA baseline, and adaptive rotation}

SOMA maintains a population of candidate points and improves them by migration toward selected targets rather than by a conventional mutation--crossover generation \citep{zelinka2004soma}. Let $x\in\mathbb{R}^D$ be a migrant and $L\in\mathbb{R}^D$ the target selected by a particular SOMA strategy. A standard continuous SOMA path can be written as
\begin{equation}
    z_j
    =
    x+t_j M_j(L-x),
    \qquad
    t_j=j\,\mathrm{Step},
    \label{eq:canonical_soma}
\end{equation}
for path positions $t_j$ up to the prescribed PathLength. Here,
\begin{equation}
    M_j=\operatorname{diag}(b_{j,1},\ldots,b_{j,D}),
\end{equation}
is a diagonal coordinate-selection matrix representing the usual PRT vector. In the independent-mask interpretation,
\begin{equation}
    b_{j,d}\sim\operatorname{Bernoulli}(p_{\mathrm{PRT}}),
\end{equation}
so $p_{\mathrm{PRT}}$ is the coordinate-activation probability. SOMA strategies can differ in target construction, whether the mask is reused or regenerated along the path, how the best path point is retained, and how boundary repair is performed; Eq.~\eqref{eq:canonical_soma} is the common leader-directed proposal underlying the present analysis.

The same rule admits a useful geometric interpretation \citep{novak2026operators}. Define the leader-relative displacement $e=x-L$. Then
\begin{equation}
    z_j-L=(I-t_jM_j)e.
    \label{eq:soma_relative}
\end{equation}
An active coordinate is therefore multiplied by $1-t_j$: $0<t_j<1$ interpolates toward the target, $t_j=1$ projects onto the target coordinate, and $t_j>1$ overshoots it. Under an independent Bernoulli mask with activation probability $p_{\mathrm{PRT}}$, the proposal moments are
\begin{align}
    \mathbb{E}[z_j\mid x,L,t_j]
    &=x+p_{\mathrm{PRT}}\,t_j(L-x),\\
    \operatorname{Cov}(z_j\mid x,L,t_j)
    &=p_{\mathrm{PRT}}(1-p_{\mathrm{PRT}})t_j^2
      \operatorname{diag}\!\bigl((L-x)_1^2,\ldots,(L-x)_D^2\bigr).
    \label{eq:soma_mask_cov}
\end{align}
These moments give a probabilistic description of the usual PRT vector and make its geometry explicit: the mask-induced covariance is diagonal in the current coordinate system. iSOMA-AR targets this axis dependence.

\paragraph{iSOMA baseline used in this study}
The baseline is the supplied improved SOMA implementation. The population size is 50. At each migration, $m=10$ individuals are sampled and the best $n=5$ of them become migrants. For each migrant, $k=15$ leader candidates are sampled and the best non-self candidate is used as $L$. Ten path locations are generated with $\mathrm{Step}=0.3$, but they are evaluated in reverse order,
\begin{equation}
    t\in\{3.0,2.7,\ldots,0.3\},
\end{equation}
so the largest overshoot is tested first. A fresh PRT mask is generated at each proposal and is allowed to be all zero. Its activation probability increases with consumed evaluation budget,
\begin{equation}
    p_{\mathrm{PRT}}
    =
    0.1+0.9\,\frac{N_{\mathrm{FE}}}{B},
    \label{eq:prt_schedule}
\end{equation}
where $B$ is the FE budget. Proposals are clipped to the box constraints and the path terminates at the first point satisfying $f(z)\leq f(x)$; unevaluated inner path points consume no FEs. After more than $50$ population-size-equivalent unsuccessful migrant attempts, 10\% of the population is uniformly reinitialized if the global best has not improved. Thus, the baseline combines rank-based migrant selection, tournament-style target selection, reverse-path overshoot, first-improvement acceptance, adaptive PRT, and a stagnation restart.

\paragraph{iSOMA-AR: Adaptive-rotation extension}
iSOMA-AR leaves all of the preceding selection, path, repair, and acceptance rules unchanged. It learns only a coordinate system in which the Bernoulli mask can occasionally be applied. 

\begin{figure}[htpb]
    \centering
    \includegraphics[width=0.85\linewidth]{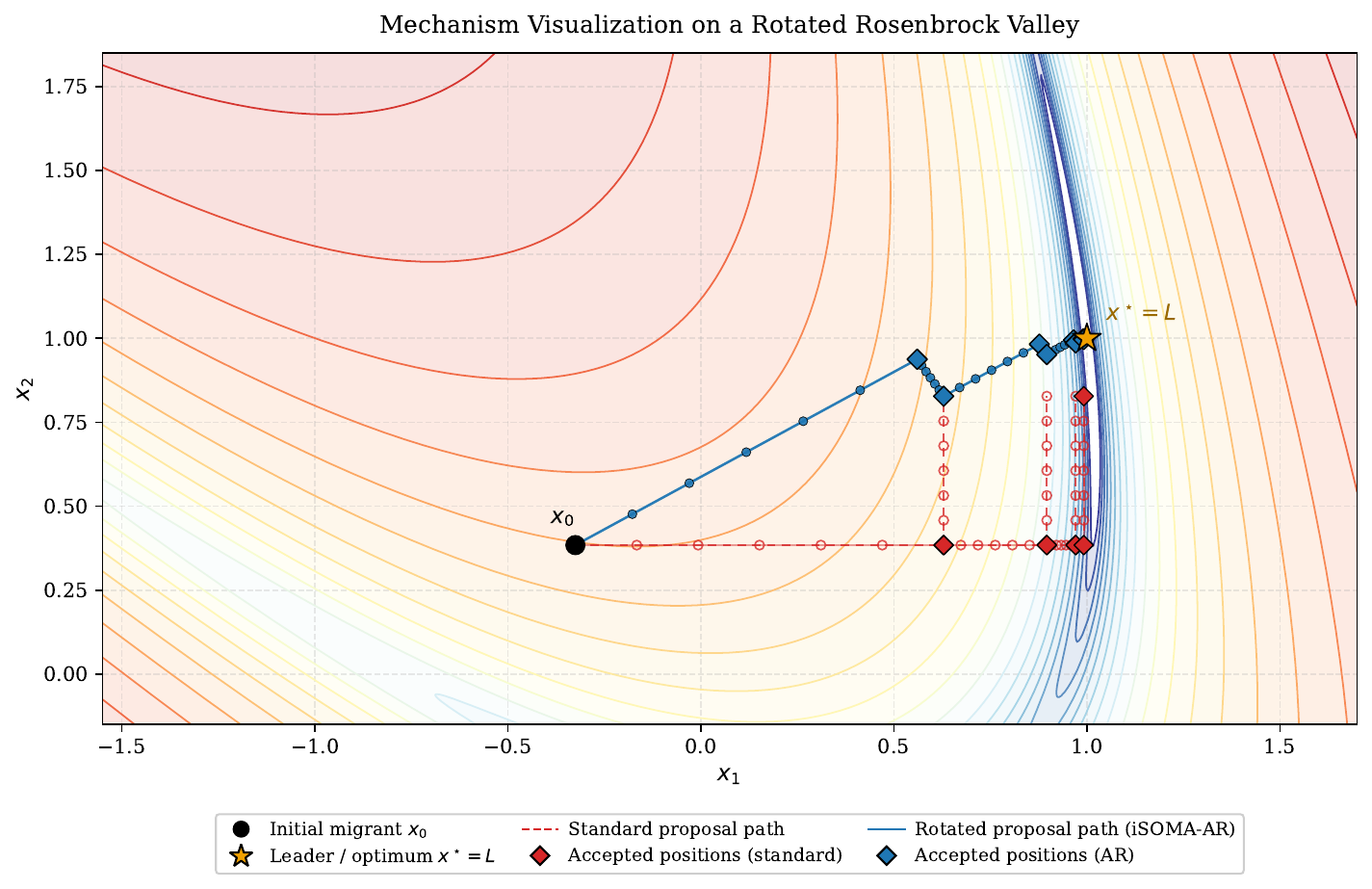}
    \caption{Illustrative proposal geometry on a rotated Rosenbrock valley. With the same partially active perturbation mask, standard iSOMA applies masking in the original coordinate system, whereas iSOMA-AR applies it in the learned basis \(Q_s\). The rotated proposals follow the valley geometry more directly and make faster progress toward the leader. Discrete points denote evaluated path locations and diamonds denote accepted replacements.}
    \label{fig:ar_geometry}
\end{figure}

For every strictly improving accepted displacement $s=z-x$, define $u=s/\lVert s\rVert$. Starting from $C_s=I$, the success matrix is updated as
\begin{equation}
    C_s\leftarrow(1-\eta)C_s+\eta uu^\top,
    \label{eq:csucc}
\end{equation}
with
\begin{equation}
    \eta
    =
    \min\!\left[
        0.35,\,
        0.12\left(
            1+\min\!\left\{2,\frac{f(x)-f(z)}{|f(x)|+\varepsilon}\right\}
        \right)
    \right].
\end{equation}
After every five successful updates, the eigenvectors of $C_s+10^{-4}I$ form an orthogonal basis $Q_s$. Rotation remains disabled until at least
$\max(6,\lceil1.5D\rceil)$ successful displacements have been observed and the off-axis score
\begin{equation}
    \rho=
    \frac{\lVert C_s-\operatorname{diag}(C_s)\rVert_F}
         {\lVert C_s\rVert_F+\varepsilon}
\end{equation}
exceeds 0.18. When this gate is open, the probability of using the rotated mask is
\begin{equation}
    p_{\mathrm{rot}}
    =
    \min\!\left[
        0.45,\,
        0.45\,\frac{\rho-0.18}{1-0.18}
    \right].
\end{equation}

A rotated proposal replaces the coordinate-masked displacement in Eq.~\eqref{eq:canonical_soma} by
\begin{equation}
    d_{\mathrm{AR}}
    =
    Q_s M Q_s^\top(L-x),
    \label{eq:ar}
\end{equation}
where $M=\operatorname{diag}(b_1,\ldots,b_D)$ is a Bernoulli mask generated with the same coordinate-activation probability $p_{\mathrm{PRT}}$. Thus, $Q_s M Q_s^\top$ projects the leader-directed displacement onto the randomly selected eigenvector directions of $C_s$. Otherwise the ordinary coordinate mask is used. The success matrix and gate are reset by the same stagnation restart as the baseline. No extra objective evaluations are introduced: iSOMA-AR changes proposal geometry only.

Figure~\ref{fig:ar_geometry} illustrates the geometric effect of Eq.~\eqref{eq:ar} on a rotated, ill-conditioned Rosenbrock valley. Under ordinary masking, the active coordinates are tied to the original axes, so successive proposals can cut across the narrow valley and make inefficient progress toward the target. When rotation is active, the same Bernoulli mask is instead applied in the learned basis \(Q_s\), allowing the proposal direction to align with correlated successful displacements and therefore with the local valley geometry.

\subsection{Comparator algorithms and common evaluation protocol}

The primary BBOB and CEC 2011 panel contains five methods: iSOMA, iSOMA-AR, jSO-derived, iL-SHADE, and CMA-ES. iL-SHADE is run through PyADE and CMA-ES through \texttt{pycma}. The custom jSO-derived implementation is retained as the primary comparator because it was used in the original experiment. A paper-faithful canonical jSO implementation is rerun on the same BBOB and CEC 2011 grids, budgets, and seed labels as a post-hoc sensitivity analysis. Both jSO variants are shown in the revised BBOB/CEC figures, while the original five-method ranks remain the primary analysis. Appendix~\ref{app:optimizers} gives the full settings and implementation differences.

The noisy VQE experiment adds SPSA, while the exact-objective VQE archive also contains auxiliary BFGS, SciPy-DE, and L-SRTDE runs. These auxiliary methods are retained for completeness, but the headline cross-benchmark comparisons emphasize the methods shared with the BBOB/CEC panel plus SPSA where noise is present.

All methods are charged against a strict objective-function evaluation counter. Within each benchmark condition and stochastic repetition, every optimizer receives the same integer seed label; heterogeneous algorithms still consume different random streams. Full implementation settings, initialization rules, repair operators, and stopping conditions are given in Appendix~\ref{app:optimizers}.

\begin{table}[htpb]
\centering
\caption{Benchmark protocols used in the final study.}
\label{tab:protocol}
\scriptsize
\begin{tabular}{@{}lll@{}}
\toprule
 & BBOB & CEC 2011 real-world \\
\midrule
Problems & $24$ noiseless functions & $22$ real-world problems \\
Dimensions & $D\in\{5,10,20\}$ & problem-specific ($1$--$240$) \\
Instances & $1$--$5$ & fixed problem definitions \\
FE budgets & $500D,\ 2000D$ & $50{,}000$ \\
Stochastic runs & $10$ / condition & $10$ / problem \\
Primary optimizers & \multicolumn{2}{l}{iSOMA, iSOMA-AR, jSO-derived, iL-SHADE, CMA-ES} \\
Post-hoc sensitivity & \multicolumn{2}{l}{jSO-canonical} \\
Primary recorded runs & $36{,}000$ & $1{,}100$ \\
Additional canonical-jSO runs & $7{,}200$ & $220$ \\
\bottomrule
\end{tabular}
\end{table}

\subsection{BBOB experiment}

The complete 24-function noiseless BBOB suite was used \citep{finck2009bbob}. The benchmark was evaluated at $D\in\{5,10,20\}$, instances 1--5, and budgets of $500D$ and $2000D$ evaluations. Ten independent stochastic repetitions were performed for every optimizer and condition, yielding 36,000 optimizer runs. The objective functions were instantiated through \texttt{cma.bbobbenchmarks}; the BBOB domain was $[-5,5]^D$.

Final-error analysis used the median over the ten stochastic repetitions for every function--dimension--instance--budget condition. Algorithms were ranked within each matched condition only; raw objective errors were not averaged across functions with incompatible scales. Pairwise summaries report wins/ties/losses (W/T/L) and the median ratio of condition-median errors.

For fixed-target analysis we followed the COCO/BBOB principle of assessing the runtime required to reach multiple target precisions \citep{hansen2021coco,hansen2016performance}. Exact first-hit evaluation counts were stored for 51 targets,
\[
\Delta f \in \{10^2,\ldots,10^{-8}\},
\]
logarithmically spaced. Empirical cumulative distribution functions (ECDFs) use evaluations divided by dimension and are reported at the largest budget, $2000D$. These are COCO-style analyses generated from exact histories, although the experimental driver was not the official COCO logging stack.

The five official BBOB groups were used as defined by the suite: separable ($f_1$--$f_5$), low/moderate conditioning ($f_6$--$f_9$), high-conditioning unimodal ($f_{10}$--$f_{14}$), multimodal with adequate global structure ($f_{15}$--$f_{19}$), and multimodal with weak global structure ($f_{20}$--$f_{24}$). We additionally use overlapping diagnostic groups for anisotropy, valley/ridge following, structured multimodality, weak-structure basin selection, ruggedness, asymmetry/deception, and boundary-oriented search. The latter are interpretive rather than an official BBOB taxonomy.

\subsection{CEC 2011 real-world experiment}

Transfer to application-oriented objectives was studied on all 22 CEC 2011 real-world optimization problems \citep{das2010cec2011}. Problem-specific dimensions and bounds were obtained from the Minion/MinionPy implementation \citep{muzakka2025minion}. Each optimizer received 50,000 evaluations and was repeated ten times, yielding 1,100 runs. The same optimizer implementations used in BBOB were imported directly into the CEC driver.

The CEC problems have heterogeneous objective scales and, for some problems, no universally usable target value in the recorded backend. Final comparisons use the median objective value per problem. For a scale-normalized convergence view, a diagnostic target ECDF was reconstructed from saved checkpoints using the backend-provided optimum when finite and otherwise the best observed value, with six relative reductions of a common starting-gap scale ($10^{-1}$ to $10^{-6}$). This ECDF is used only as a diagnostic outside the official CEC/COCO metrics.

For interpretation, the 22 problems were grouped into five application families: control/parameter estimation, molecular/material potentials, signal/antenna design, power systems, and spacecraft trajectories. These groups are interpretive and outside the official CEC taxonomy.

The Minion CEC 2011 rewrite has documented discrepancies relative to the Octave/MATLAB reference on F3, F4, F21, and F22. Consequently, results involving those four problems are marked as provisional and should be spot-validated with the reference implementation before final publication.

\subsection{Variational quantum benchmark and sampling noise}

\paragraph{Hamiltonian family and fixed disorder instances.}
All quantum experiments minimize the variational energy
\begin{equation}
    E(\theta;\Gamma,\lambda)
    =
    \langle\psi(\theta)|H(\Gamma,\lambda)|\psi(\theta)\rangle,
    \qquad
    |\psi(\theta)\rangle=U(\theta)|0\rangle .
    \label{eq:vqe_energy}
\end{equation}
The three benchmark models share a sparse degree-three graph $\mathcal{E}$ formed by a nearest-neighbour ring and an opposite-site matching edge for every qubit pair $i$ and $i+N/2$. Their common frustrated Ising backbone is
\begin{equation}
    H_{\mathrm{SG}}
    =
    \sum_{(i,j)\in\mathcal E}J^z_{ij}Z_iZ_j
    +
    \sum_{i=1}^{N}h_i Z_i ,
    \label{eq:hsg}
\end{equation}
with $J^z_{ij}\in\{-1,+1\}$ and $h_i\sim U[-0.2,0.2]$. The full two-parameter family is
\begin{equation}
    H(\Gamma,\lambda)
    =
    H_{\mathrm{SG}}
    -
    \Gamma\sum_{i=1}^{N}X_i
    +
    \lambda\sum_{(i,j)\in\mathcal E}
        \left(J^x_{ij}X_iX_j+J^y_{ij}Y_iY_j\right).
    \label{eq:unified_h}
\end{equation}
The three labels used throughout this paper are therefore defined internally as
\begin{align}
    \mathrm{Q1}:&\quad H(0,0)=H_{\mathrm{SG}},\\
    \mathrm{Q2}:&\quad H(0.3,0)=H_{\mathrm{SG}}-0.3\sum_i X_i,\\
    \mathrm{Q3}:&\quad H(0,0.5)
       =H_{\mathrm{SG}}
       +0.5\sum_{(i,j)\in\mathcal E}
        \left(J^x_{ij}X_iX_j+J^y_{ij}Y_iY_j\right).
\end{align}

For every Ising backbone required by Q1/Q2, 50 candidate $\{J^z,h\}$ realizations are generated from a fixed master seed and the instance with the largest exact number of one-spin-flip local minima is retained. The same saved $J^z,h$ backbone is reused when moving from Q1 to Q2, so the transverse field is the only Hamiltonian change at a matched size. For Q3 at $N=10$ and $N=12$, the same paired $J^z,h$ data are again reused, while $J^x$ and $J^y$ are fixed independent $\pm1$ draws from seed 1234. Q3 at $N=8$ uses the corresponding fully fixed seed-1234 realization. No disorder realization is regenerated between optimizers, budgets, or noise levels. Additional construction and reference details are given in Appendix~\ref{app:vqe_details}.

\paragraph{Variational ansatz and error definitions.}
For Q1 we use the product-state circuit
\begin{equation}
    U_{\mathrm{prod}}(\theta)
    =
    \bigotimes_{i=1}^{N}R_y(\theta_i),
    \qquad D=N,
\end{equation}
with $\theta_i\in[-\pi,\pi]$. Because $\theta_i\in\{0,\pi\}$ spans every computational-basis configuration, the exact diagonal ground state is variationally reachable. The objective also has the analytic form
\begin{equation}
    E_{\mathrm{Q1}}(\theta)
    =
    \sum_{(i,j)\in\mathcal E}
      J^z_{ij}\cos\theta_i\cos\theta_j
    +
    \sum_i h_i\cos\theta_i .
    \label{eq:q1_analytic}
\end{equation}

Q2 and Q3 use the same shallow $2N$-parameter hardware-efficient circuit,
\begin{equation}
    U_{\mathrm{HEA}}(\theta)
    =
    \left[\bigotimes_{i=1}^{N}R_y(\theta_{N+i})\right]
    U_{\mathrm{ent}}
    \left[\bigotimes_{i=1}^{N}R_y(\theta_i)\right],
    \qquad D=2N,
\end{equation}
where
\begin{equation}
    U_{\mathrm{ent}}
    =
    \mathrm{CNOT}_{N-1\rightarrow N}\cdots
    \mathrm{CNOT}_{2\rightarrow3}
    \mathrm{CNOT}_{1\rightarrow2}.
\end{equation}
All quantum objectives are evaluated by exact NumPy statevector arithmetic in the noiseless experiment. Q1 and Q2 are scored by physical ground-state error $E_{\mathrm{best}}-E_0$, with $E_0$ obtained by exact enumeration or sparse diagonalization. Because the fixed Q3 circuit need not contain the physical ground state, Q3 is scored by optimization error
$E_{\mathrm{best}}-E_{\mathrm{ref}}$ relative to a fixed best-known variational reference. The Q3 reference construction is defined in Appendix~\ref{app:vqe_details}; iSOMA-AR did not improve any of those final reference values.

\paragraph{Exact-objective protocol.}
Q1 uses $N\in\{12,14,16\}$, Q2 uses $N\in\{10,12\}$, and Q3 uses $N\in\{8,10,12\}$. Every size is evaluated at 10,000 and 30,000 FEs with 25 independent seed labels, 420--444. The exact comparison dataset contains 25 runs of each reference optimizer in every condition, and iSOMA-AR is evaluated on the identical saved instances, bounds, budgets, and seed set. This permits direct AR--baseline comparisons without changing the physical problem or evaluation allowance.

\paragraph{Sampling-noise protocol.}
A separate robustness experiment uses the common $N=12$ condition for Q1--Q3, the same 10,000/30,000-FE budgets, and the same 25 seed labels. The compared methods are iSOMA, iSOMA-AR, jSO-derived, iL-SHADE, CMA-ES, and SPSA. Every FE returned to the optimizer is corrupted independently,
\begin{equation}
    \widetilde E(\theta)
    =
    E(\theta)+\epsilon,
    \qquad
    \epsilon\sim\mathcal N(0,\sigma_E^2).
    \label{eq:noise}
\end{equation}
Two effective shot counts are used: $M=8192$ (low noise) and $M=128$ (high noise). We set
\begin{equation}
    \sigma_{\mathrm{norm}}=\frac{0.5}{\sqrt{M}},
    \qquad
    \sigma_E=(E_{\max}-E_{\min})\sigma_{\mathrm{norm}},
    \label{eq:noise_scale}
\end{equation}
so the dimensionless noise level is shared while the optimizer still receives raw Hamiltonian energies. We refer to this approximation as an effective Gaussian sampling-noise surrogate; it omits Pauli-term-resolved measurement statistics.

Every queried parameter vector is also evaluated by an exact oracle for offline scoring; the optimizer receives only the noisy value. The primary noisy endpoint is the \emph{exact} energy of the parameter vector that produced the smallest noisy observation. A secondary oracle-best-seen diagnostic records the lowest exact energy among all queried points in the same run and quantifies selection loss. Q1 uses Eq.~\eqref{eq:q1_analytic}; Q2 uses a vectorized NumPy $R_y$--CNOT--$R_y$ statevector evaluator with diagonal and $X$ expectations; Q3 uses the corresponding vectorized Heisenberg-statevector evaluator.

\subsection{Statistical summaries}

For BBOB and CEC 2011, final-condition medians were compared by paired W/T/L counts and paired Wilcoxon signed-rank tests on log-scaled error or regret ratios, with Holm correction over the reported pairwise comparisons. We use these $p$-values descriptively because dimensions, budgets, and instances derived from the same benchmark functions are not fully independent experimental units.

The VQE study contains 25 independent optimizer runs per condition. Shared integer seed labels serve as reproducibility indices rather than common random numbers because the algorithms consume randomness differently. Each condition is analyzed with a Kruskal--Wallis omnibus test followed by two-sided Mann--Whitney $U$ comparisons with Holm correction; rank-biserial effects are reported in the appendix. Main-text interpretation emphasizes medians, interquartile ranges, condition-wise ranks, and direct iSOMA-AR versus iSOMA comparisons. Wall-clock time per 1000 objective evaluations is included as an implementation-overhead indicator.

\subsection{Targeted mechanism and implementation controls}

The adaptive-rotation mechanism was probed with a targeted paired ablation instead of a second full benchmark. The controlled part uses $D=10$ axis-aligned and orthogonally rotated Ellipsoid and Rastrigin landscapes, 20,000 FEs, and 25 paired runs. Default learned AR is compared with baseline iSOMA, a random-basis control that preserves the same learned gate probability while replacing the learned orientation by a fixed random orthogonal basis, two gate thresholds ($0.10$ and $0.30$ versus the default $0.18$), and two maximum rotation probabilities ($0.25$ and $0.65$ versus the default $0.45$). A small BBOB corroboration uses only $f_2$, $f_{10}$, $f_3$, $f_{15}$, $f_8$, and $f_9$ at $D=10$, instances 1--2, and 2000 FE/D.

The all-zero-PRT control uses Q1 and Q2 at $N=12$ under the high-noise surrogate ($M=128$ effective shots), with the same 10,000/30,000-FE budgets and 25 seed labels as the main noise study. For both iSOMA and iSOMA-AR, the original independently sampled Bernoulli mask is compared with a minimally repaired version in which an all-zero mask activates one uniformly random coordinate. Paired variants share the same FE-indexed Gaussian-noise stream. This isolates the mask rule without changing the remaining migration mechanism.

Canonical jSO is rerun on the complete BBOB and CEC 2011 grids using the paper-faithful implementation described in Appendix~\ref{app:optimizers}. The VQE study retains the original jSO-derived comparator.

\section{Results}

\subsection{BBOB benchmark}

Table~\ref{tab:bbob_overall} summarizes the full five-instance BBOB experiment. The jSO-derived comparator obtained the lowest mean rank (1.86), followed by CMA-ES (2.49), iL-SHADE (3.13), iSOMA-AR (3.46), and baseline iSOMA (4.05). The ordering by winner frequency is slightly different: CMA-ES attained the lowest condition-median error in 51.0\% of matched conditions and jSO-derived in 49.9\%. Their direct final-error comparison was nearly balanced (310/115/295 for jSO-derived/CMA-ES) and showed no overall paired difference after Holm correction ($p_{\mathrm{Holm}}=0.99$), indicating complementary rather than uniformly ordered performance profiles.

Adaptive rotation produced a clear BBOB improvement over the baseline. iSOMA-AR beat iSOMA in 507 of 720 matched conditions, tied in 42, and lost in 171. The median condition-wise error ratio was approximately 0.81 and the paired comparison was strongly shifted toward iSOMA-AR ($p_{\mathrm{Holm}}<10^{-49}$). The improvement was obtained with limited computational overhead: median time per 1000 evaluations increased by about 12\% relative to baseline iSOMA.

\begin{table*}[htpb]
\centering
\caption{Overall BBOB results across 720 matched function--dimension--instance--budget conditions. Target-hit fraction is the fraction of all run--target pairs attained among the 51 fixed targets $\Delta f\in[10^2,10^{-8}]$ at the $2000D$ budget. Lower rank and runtime are better; higher shares are better.}
\label{tab:bbob_overall}
\small
\begin{tabular}{@{}lrrrrrr@{}}
\toprule
Algorithm & Mean rank & Top-1 & Target-hit frac. & Success $10^{-8}$ & Med. FE/D to $10^{-8}$ & s/1000 FE \\
\midrule
jSO-derived & 1.86 & 49.9\% & 65.4\% & 48.9\% & 1073 & 0.0279 \\
CMA-ES & 2.49 & 51.0\% & 55.1\% & 44.4\% & 417 & 0.0676 \\
iL-SHADE & 3.13 & 16.9\% & 35.8\% & 15.0\% & 552 & 0.0334 \\
iSOMA-AR & 3.46 & 10.6\% & 36.4\% & 15.5\% & 1212 & 0.0204 \\
iSOMA & 4.05 & 8.3\% & 29.5\% & 12.2\% & 1154 & 0.0182 \\
\bottomrule
\end{tabular}
\end{table*}

\begin{figure}[htpb]
    \centering
    \includegraphics[width=0.75\linewidth]{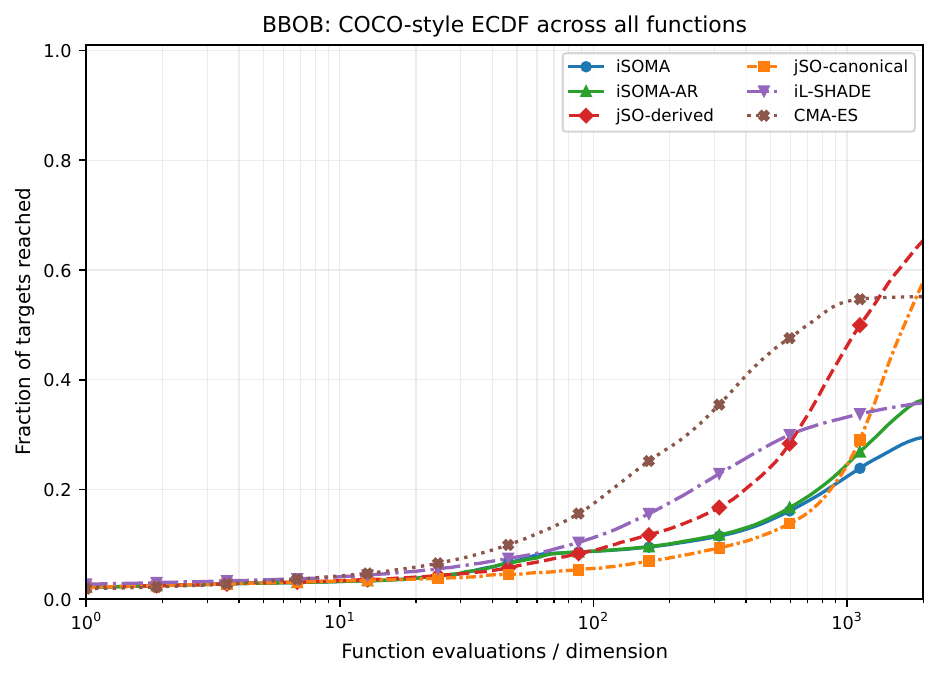}
    \caption{COCO-style fixed-target ECDF across all BBOB functions, dimensions, instances, and 51 target precisions at the $2000D$ budget. The jSO-derived and jSO-canonical implementations are shown as separate curves; canonical jSO is a post-hoc sensitivity comparator and is not included in the primary five-method rank analysis.}
    \label{fig:bbob_ecdf_overall}
\end{figure}

The exact fixed-target view in Fig.~\ref{fig:bbob_ecdf_overall} supports the same conclusion while distinguishing target coverage from convergence speed. Across all 51 run--target pairs, the fraction attained by $2000D$ increased from 29.5\% for iSOMA to 36.4\% for iSOMA-AR. At $\Delta f=10^{-8}$, the success fraction increased from 12.2\% to 15.5\%. iSOMA-AR and iL-SHADE had similar tight-target success fractions (15.5\% and 15.0\%, respectively), but iL-SHADE reached successful $10^{-8}$ targets earlier in the median (552 FE/D versus 1212 FE/D). CMA-ES showed the fastest successful target attainment, whereas jSO-derived reached the largest fraction of targets overall.

\paragraph{Landscape dependence}

\begin{figure*}[htpb]
    \centering
    \includegraphics[width=\textwidth]{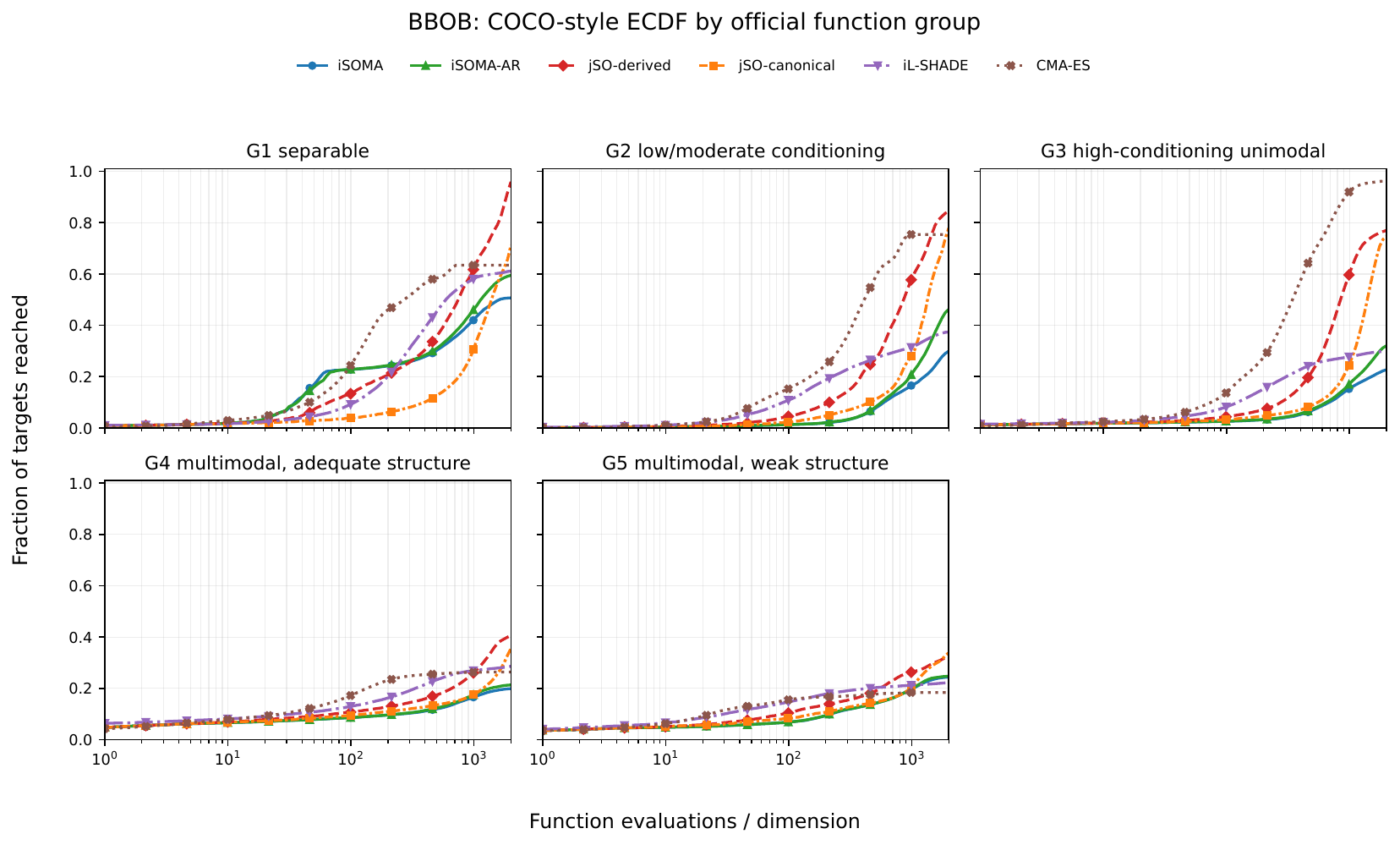}
    \caption{COCO-style fixed-target ECDF split by the five official BBOB function groups.}
    \label{fig:bbob_ecdf_groups}
\end{figure*}

The official BBOB groups reveal where the adaptive basis helps and where it remains insufficient (Fig.~\ref{fig:bbob_ecdf_groups} and Table~\ref{tab:bbob_groups}). Relative to baseline iSOMA, the strongest median error reductions occurred in low/moderate conditioning (AR/base ratio 0.55) and high-conditioning unimodal functions (0.45). iSOMA-AR won 137 of 150 matched conditions against the baseline in the high-conditioning group. This is direct evidence that the learned basis corrects an important part of the coordinate sensitivity that motivated the modification.

The gain over baseline does not close the gap to the strongest comparators on these landscapes. In the high-conditioning unimodal group, iSOMA-AR lost 105 of 150 conditions to iL-SHADE, while CMA-ES was the most frequent condition winner in 131 of 150 cases. A similar gap remained on multimodal functions with adequate global structure: iSOMA-AR improved on iSOMA in 100 of 150 conditions but lost 129 of 150 to iL-SHADE. By contrast, weak-global-structure multimodality was much closer: AR/iL-SHADE was 69/3/78 with a median error ratio near one.

\begin{table*}[htpb]
\centering
\caption{BBOB official-group pairwise final-error outcomes. W/T/L is the number of matched conditions won/tied/lost by the first method. The ratio is the median error ratio; values below one favor iSOMA-AR.}
\label{tab:bbob_groups}
\small
\resizebox{\textwidth}{!}{%
\begin{tabular}{@{}lp{3.0cm}rrrr@{}}
\toprule
Group & Most frequent condition winner & AR/base W/T/L & AR/base ratio & AR/iL-SHADE W/T/L & AR/iL-SHADE ratio \\
\midrule
G1 separable & jSO-derived 115/150 & 84/39/27 & 0.93 & 71/33/46 & 1.00 \\
G2 low/moderate conditioning & CMA-ES 90/120 & 99/0/21 & 0.55 & 45/0/75 & 1.32 \\
G3 high-conditioning unimodal & CMA-ES 131/150 & 137/0/13 & 0.45 & 45/0/105 & 2.53 \\
G4 multimodal, adequate structure & jSO-derived 80/150 & 100/0/50 & 0.90 & 21/0/129 & 2.44 \\
G5 multimodal, weak structure & jSO-derived 67/150 & 87/3/60 & 0.99 & 69/3/78 & 1.01 \\
\bottomrule
\end{tabular}
}
\end{table*}

\begin{figure*}[htpb]
    \centering
    \includegraphics[width=0.8\textwidth]{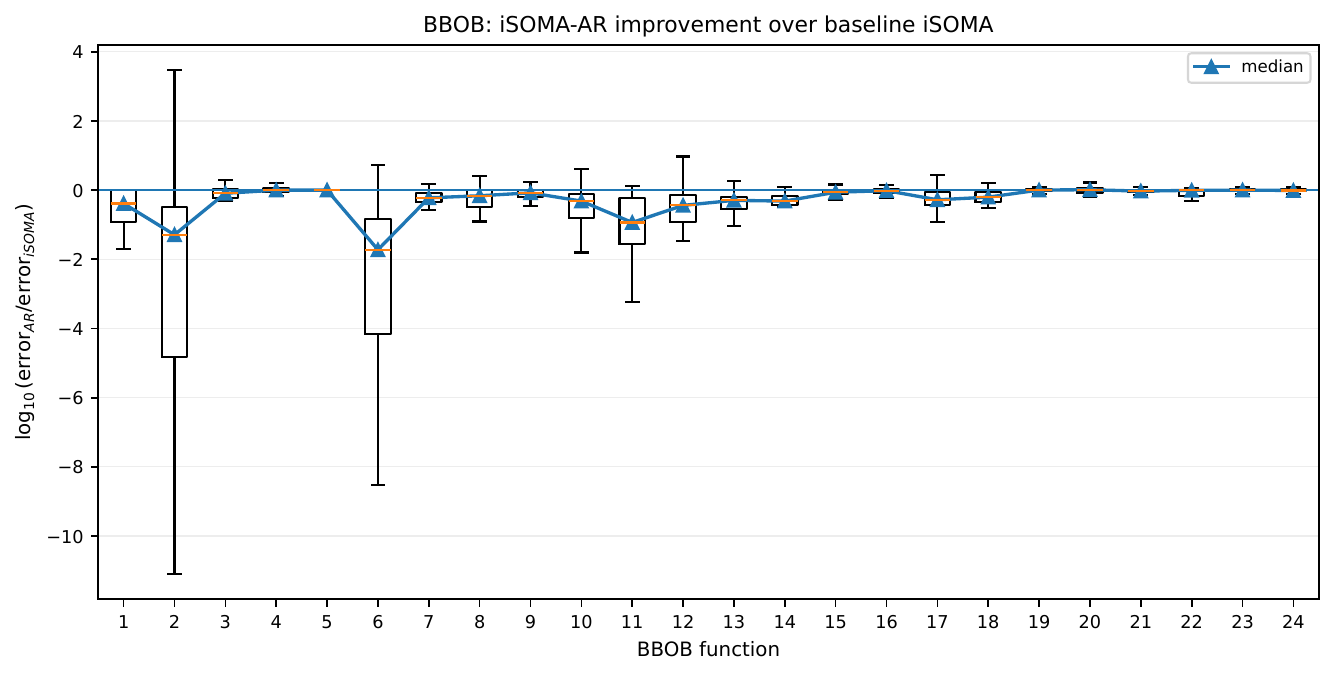}
    \caption{Distribution of $\log_{10}(\mathrm{error}_{\mathrm{AR}}/\mathrm{error}_{\mathrm{iSOMA}})$ by BBOB function. Values below zero favor iSOMA-AR.}
    \label{fig:bbob_ar_function}
\end{figure*}

The function-level pattern is more specific (Fig.~\ref{fig:bbob_ar_function}). On separable Rastrigin $f_3$, iSOMA-AR beat iL-SHADE in 29/30 matched conditions; on asymmetric B\"uche--Rastrigin $f_4$, it won all 30. It also beat iL-SHADE on all 30 Schwefel $f_{20}$ conditions and on 21/30 Gallagher-101 $f_{21}$ conditions. These cases show that the method can be competitive when multimodality, asymmetry, boundary interaction, or weak global structure dominate.

The opposite pattern appears on strongly oriented or fine-scale rugged landscapes. On the rotated ellipsoid $f_{10}$, iSOMA-AR improved on baseline in 28/30 conditions but lost to iL-SHADE in 23/30; CMA-ES won every matched condition. On Bent Cigar $f_{12}$ the corresponding AR/iL-SHADE record was 5/0/25. iL-SHADE also dominated on Weierstrass $f_{16}$, Katsuura $f_{23}$, and Lunacek bi-Rastrigin $f_{24}$. Thus, adaptive rotation fixes a substantial baseline weakness but does not by itself provide the covariance adaptation, success-history control, or basin-selection behavior needed for all difficult BBOB geometries.

\paragraph{Targeted adaptive-rotation ablation}
The controlled ablation isolates the contribution of the learned orientation. On the rotated Ellipsoid, default AR significantly outperformed both baseline iSOMA and the random-basis control ($p=1.2\times10^{-4}$ for both paired comparisons), while the Rastrigin controls showed no corresponding advantage. Selected BBOB tests show the same pattern on rotated Rosenbrock $f_9$ and rotated Ellipsoid $f_{10}$. Threshold and rotation-cap perturbations preserve the qualitative result (Appendix~\ref{app:ar_ablation}).

\paragraph{jSO-derived and canonical jSO}

The jSO-derived method deserves separate attention. It was the most frequent winner on separable multimodal functions, Step Ellipsoidal, Schaffers F7 and its ill-conditioned counterpart, and several weak-global-structure functions. CMA-ES instead dominated the smooth ill-conditioned and ridge-like functions. This explains why their overall primary-panel comparison is balanced even though their group profiles differ strongly.

Canonical jSO also performed strongly on BBOB, although jSO-derived remained better on the aggregate fixed-target and direct paired comparisons. The two implementations are reported separately because their population initialization, parameter memories, mutation weights, stage restrictions, and boundary handling differ materially. Full six-method and direct comparisons are given in Appendix~\ref{app:jso_sensitivity}.

\subsection{CEC 2011 real-world benchmark}

\paragraph{Overall results}

The CEC 2011 ordering differs from BBOB (Table~\ref{tab:cec_overall}). The jSO-derived method was clearly strongest, with mean rank 1.50 and the lowest median objective on 77.3\% of the 22 problems. CMA-ES ranked second (2.77), iSOMA-AR third (3.27), baseline iSOMA fourth (3.41), and iL-SHADE fifth (4.05).

Adaptive rotation transfers less consistently than on BBOB. Against baseline iSOMA, the problem-level record was 11/3/8 in favor of iSOMA-AR and the Holm-adjusted paired test was not significant ($p_{\mathrm{Holm}}=0.34$). Against iL-SHADE, iSOMA-AR won 14 problems, tied two, and lost six, with a median regret ratio of approximately 0.125 and $p_{\mathrm{Holm}}=0.052$. Under this 50,000-evaluation protocol, iSOMA-AR was therefore more competitive with iL-SHADE than in aggregate BBOB.

The jSO-derived method also transferred strongly: it beat CMA-ES on 17 problems, tied one, and lost four, with $p_{\mathrm{Holm}}=0.009$. This contrasts with the essentially balanced jSO-derived/CMA-ES comparison on BBOB.

Canonical jSO was rerun on the same 22 CEC 2011 problems, 50,000-FE budget, and ten-run protocol. It is shown alongside jSO-derived in the revised CEC figures; the original five-method ranks and target-reference construction are unchanged. Detailed sensitivity results are given in Appendix~\ref{app:jso_sensitivity}.

\begin{table*}[htpb]
\centering
\caption{Overall CEC 2011 results across 22 problems at 50,000 evaluations. The normalized target-hit fraction is diagnostic: it is the fraction of run--target pairs attained for six relative gap reductions $10^{-1}$--$10^{-6}$ reconstructed from saved checkpoints.}
\label{tab:cec_overall}
\small
\begin{tabular}{@{}lrrrrrr@{}}
\toprule
Algorithm & Mean rank & Top-1 & Target-hit frac. & Success $10^{-4}$ & Success $10^{-6}$ & s/1000 FE \\
\midrule
jSO-derived & 1.50 & 77.3\% & 63.1\% & 59.5\% & 29.1\% & 0.0202 \\
CMA-ES & 2.77 & 18.2\% & 48.6\% & 45.5\% & 7.7\% & 0.0834 \\
iSOMA-AR & 3.27 & 13.6\% & 42.0\% & 33.2\% & 11.8\% & 0.0193 \\
iSOMA & 3.41 & 9.1\% & 43.2\% & 34.1\% & 12.3\% & 0.0110 \\
iL-SHADE & 4.05 & 9.1\% & 18.3\% & 11.4\% & 6.8\% & 0.0224 \\
\bottomrule
\end{tabular}
\end{table*}

\begin{figure}[htpb]
    \centering
    \includegraphics[width=0.75\linewidth]{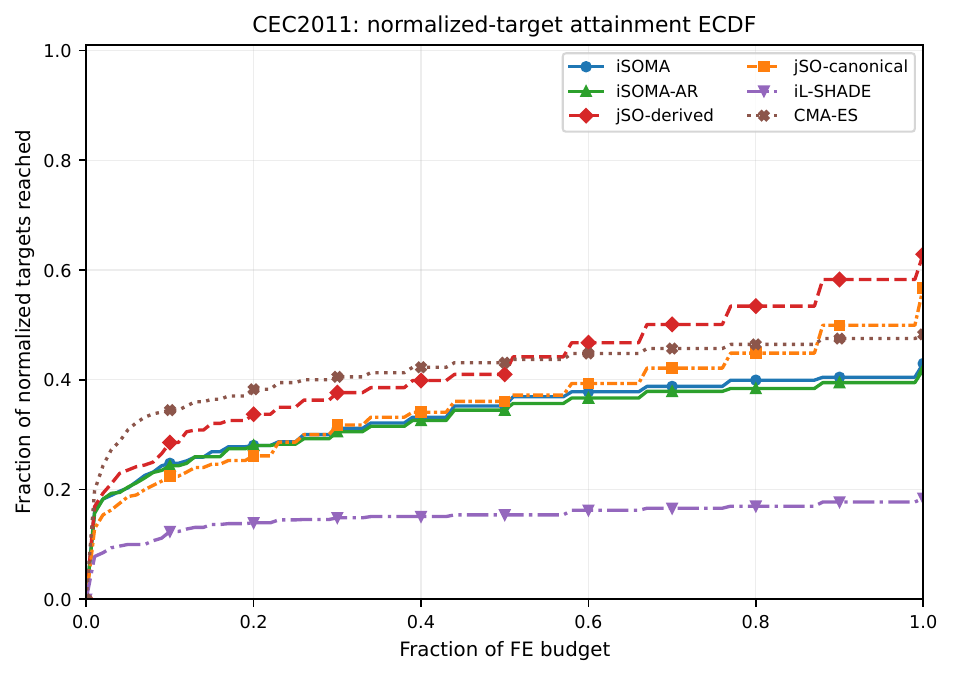}
    \caption{Diagnostic normalized-target ECDF over the 22 CEC 2011 real-world problems. Targets are relative gap reductions reconstructed from saved checkpoints; this diagnostic lies outside the official CEC/COCO metrics.}
    \label{fig:cec_ecdf}
\end{figure}

The diagnostic normalized-target ECDF (Fig.~\ref{fig:cec_ecdf}) emphasizes convergence coverage in addition to final medians. jSO-derived attained the largest fraction of target pairs (63.1\%), followed by CMA-ES (48.6\%), iSOMA (43.2\%), iSOMA-AR (42.0\%), and iL-SHADE (18.3\%). The small final-rank improvement of iSOMA-AR over baseline therefore does not translate into uniformly faster target attainment.

\paragraph{Application families}

\begin{figure*}[htpb]
    \centering
    \includegraphics[width=0.9\textwidth]{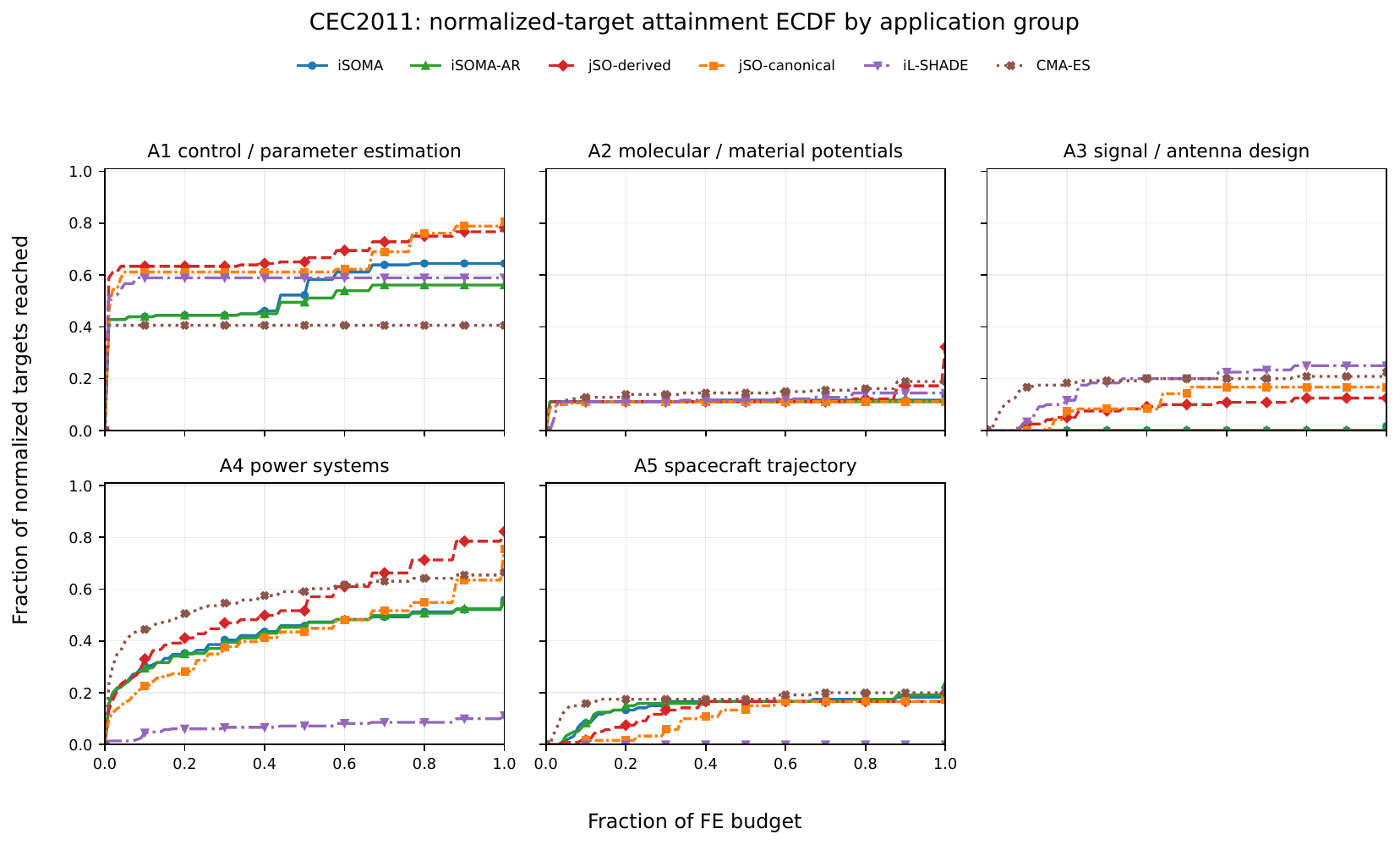}
    \caption{Diagnostic CEC 2011 target-attainment ECDF split by application family. The groups are interpretive and outside the official benchmark taxonomy.}
    \label{fig:cec_groups}
\end{figure*}

The application-family breakdown shows a clear domain dependence (Table~\ref{tab:cec_groups} and Fig.~\ref{fig:cec_groups}). The strongest result for iSOMA-AR occurred in power-system optimization. Across the 12 transmission, dispatch, and hydrothermal scheduling problems, it beat baseline iSOMA on eight, tied one, and lost three; against iL-SHADE it won 10 of 12. The jSO-derived method was nevertheless the most frequent winner, taking 10 of the 12 power-system problems.

On molecular/material potential problems, adaptive rotation was not beneficial: iSOMA-AR lost all three problem-level comparisons to baseline iSOMA and two of three to iL-SHADE. Signal/antenna design favored CMA-ES and iL-SHADE. The control/parameter-estimation set was too small and contained two effectively tied one-dimensional problems, so little should be inferred from its aggregate ordering.

F21 and F22 both have documented Minion reference discrepancies, so the two-problem spacecraft result is treated as provisional.

\begin{table*}[htpb]
\centering
\caption{CEC 2011 diagnostic application-group outcomes. The groups are interpretive and outside the official CEC 2011 taxonomy. W/T/L compares median final regret over the problems in each group; values below one favor iSOMA-AR.}
\label{tab:cec_groups}
\small
\resizebox{\textwidth}{!}{%
\begin{tabular}{@{}lp{3.0cm}rrrr@{}}
\toprule
Group & Most frequent problem winner & AR/base W/T/L & AR/base ratio & AR/iL-SHADE W/T/L & AR/iL-SHADE ratio \\
\midrule
A1 control / parameter estimation & jSO-derived 3/3 & 0/2/1 & 1 & 1/2/0 & 1 \\
A2 molecular / material potentials & jSO-derived 3/3 & 0/0/3 & 1.3 & 1/0/2 & 1.9 \\
A3 signal / antenna design & CMA-ES 1/2, iL-SHADE 1/2 & 2/0/0 & 0.97 & 0/0/2 & 4.9 \\
A4 power systems & jSO-derived 10/12 & 8/1/3 & 0.87 & 10/0/2 & 0.0002 \\
A5 spacecraft trajectory & iSOMA-AR 1/2, jSO-derived 1/2 & 1/0/1 & 0.88 & 2/0/0 & 0.038 \\
\bottomrule
\end{tabular}
}
\end{table*}

The convergence-rank trajectories in Appendix Fig.~\ref{fig:cec_convergence} further show that the relative order is not purely an endpoint effect. In particular, the advantage of jSO-derived on power-system problems develops over the evaluation budget, while the SOMA variants remain competitive on a subset of those tasks without the larger per-evaluation overhead of CMA-ES.

\begin{table}[htpb]
\centering
\caption{Selected paired comparisons. Ratios are median error/regret ratios of the first method over the second; Holm-adjusted Wilcoxon $p$-values are descriptive because benchmark conditions are not fully independent experimental units.}
\label{tab:pairwise}
\small
\begin{tabular}{@{}llrrr@{}}
\toprule
Study & Comparison & W/T/L & Median ratio & $p_{\rm Holm}$ \\
\midrule
BBOB & iSOMA-AR / iSOMA & 507/42/171 & 0.814 & 1.34e-50 \\
BBOB & iSOMA-AR / iL-SHADE & 251/36/433 & 1.35 & 8.44e-12 \\
BBOB & iSOMA-AR / CMA-ES & 198/46/476 & 2.78 & 2.34e-45 \\
BBOB & iSOMA-AR / jSO-derived & 58/54/608 & 22 & 1.91e-101 \\
BBOB & jSO-derived / CMA-ES & 310/115/295 & 1 & 0.99 \\
CEC2011 & iSOMA-AR / iSOMA & 11/3/8 & 0.99 & 0.338 \\
CEC2011 & iSOMA-AR / iL-SHADE & 14/2/6 & 0.125 & 0.0516 \\
CEC2011 & iSOMA-AR / CMA-ES & 8/1/13 & 2.16 & 0.179 \\
CEC2011 & iSOMA-AR / jSO-derived & 1/2/19 & 7.55 & 0.00101 \\
CEC2011 & jSO-derived / CMA-ES & 17/1/4 & 0.42 & 0.00916 \\
\bottomrule
\end{tabular}
\end{table}

\subsection{VQE exact-objective transfer and sampling-noise robustness}

\begin{figure*}[htpb]
    \centering
    \includegraphics[width=\textwidth]{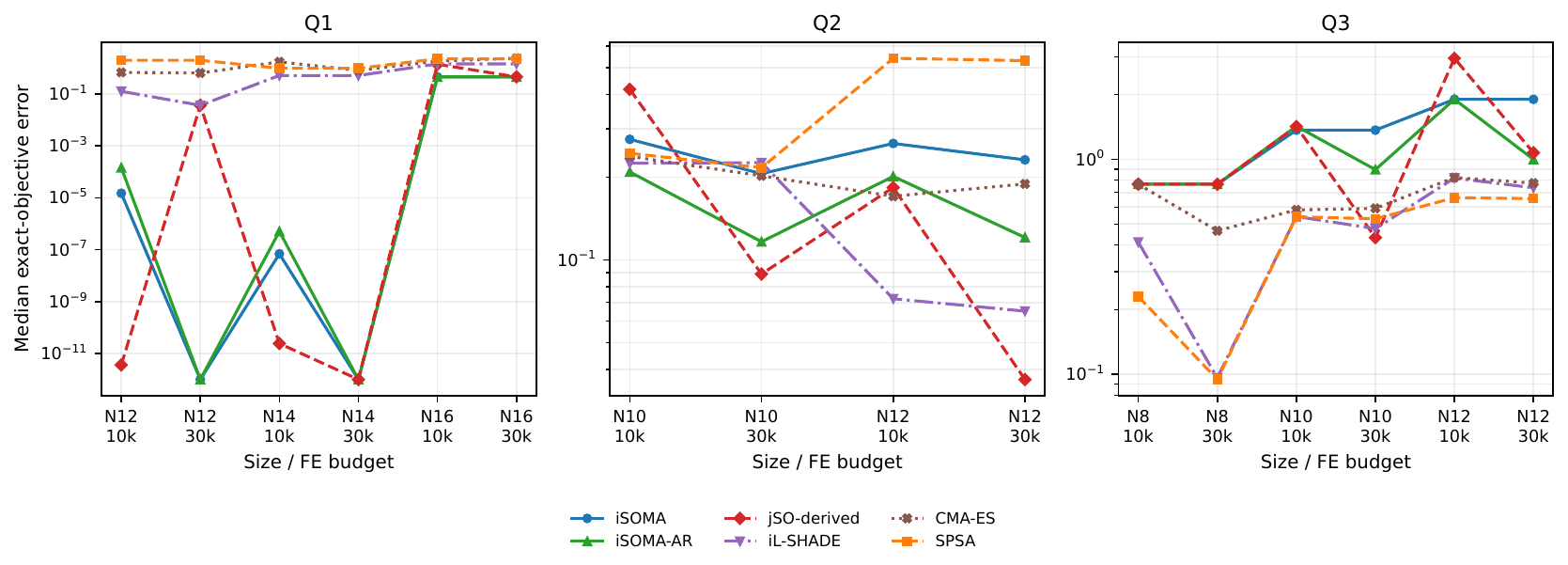}
    \caption{Exact-objective Q1--Q3 comparison using the 25-run protocol defined in Methods. The figure shows the six core methods retained for comparison with the noisy study. Q1 and Q2 use physical ground-state error; Q3 uses optimization error relative to the unchanged saved variational reference.}
    \label{fig:vqe_exact}
\end{figure*}

\paragraph{Statevector results}
The exact-objective dataset contains the complete reference-optimizer runs together with the matched iSOMA-AR runs, so the adaptive variant can be compared directly against iSOMA under identical instances, budgets, and 25-run seed labels. Table~\ref{tab:vqe_exact} and Fig.~\ref{fig:vqe_exact} show a strongly model-dependent result.

On Q1, adaptive rotation does not improve the baseline. At 10,000 FEs the iSOMA-AR median is approximately 9.7 times the iSOMA median for $N=12$ and 7.4 times the baseline for $N=14$; by 30,000 FEs both methods reach numerical zero for $N=14$ and near-zero error for $N=12$. At $N=16$ the two methods are effectively tied around 0.458. This is consistent with Q1's unusually favorable native coordinate structure: introducing a learned basis is unnecessary and can slow the sparse coordinate-hopping mechanism that already matches the landscape.

Q2 is the clearest exact-objective success of adaptive rotation. iSOMA-AR lowers the median error relative to iSOMA in all four size--budget conditions: from 0.275 to 0.209 at $N=10$/10k, 0.206 to 0.116 at $N=10$/30k, 0.265 to 0.201 at $N=12$/10k, and 0.231 to 0.121 at $N=12$/30k. These correspond to reductions of approximately 24\%, 44\%, 24\%, and 48\%, respectively. The direct AR--baseline comparison remains Holm-significant in three of the four conditions; only $N=10$/10k is not significant after correction. Across the six core methods, iSOMA-AR has the lowest mean rank on Q2 (2.50), slightly ahead of iL-SHADE and jSO-derived (2.75 each), although the best condition median still alternates between methods.

Q3 shows a weaker, budget-dependent effect. At 10,000 FEs the adaptive variant is essentially unchanged at $N=8$ and $N=12$ and slightly worse at $N=10$. At 30,000 FEs, however, its median error falls from 1.367 to 0.895 at $N=10$ and from 1.902 to 0.999 at $N=12$, reductions of approximately 35\% and 47\%. These late-budget improvements do not remain significant after Holm correction and do not make iSOMA-AR the leading Q3 method; SPSA, iL-SHADE, CMA-ES, or jSO-derived still obtain lower medians depending on the condition. The saved Q3 variational reference is unchanged, so the improvement reflects optimizer behavior rather than a shifted reference.

\begin{table*}[htpb]
\centering
\small
\caption{Exact-objective VQE extension using the original 25-run Q1--Q3 benchmark protocol. Values are median final errors. For Q1 and Q2 the error is relative to the exact physical ground energy; for Q3 it is relative to the unchanged saved best-known variational reference. AR/base is the ratio of iSOMA-AR to iSOMA median error; values below one favor adaptive rotation.}
\label{tab:vqe_exact}
\begin{tabular}{@{}llrrrl@{}}
\toprule
Model & Condition & iSOMA & iSOMA-AR & AR/base & Best core method \\
\midrule
Q1 & $N=12$, 10k & \num{1.50e-05} & \num{1.46e-04} & 9.708 & jSO-derived \\
Q1 & $N=12$, 30k & 0 & \num{1.17e-13} & -- & iSOMA \\
Q1 & $N=14$, 10k & \num{6.93e-08} & \num{5.10e-07} & 7.360 & jSO-derived \\
Q1 & $N=14$, 30k & 0 & 0 & -- & iSOMA \\
Q1 & $N=16$, 10k & 0.458 & 0.458 & 1.000 & iSOMA \\
Q1 & $N=16$, 30k & 0.458 & 0.458 & 1.000 & jSO-derived \\
Q2 & $N=10$, 10k & 0.275 & 0.209 & 0.761 & iSOMA-AR \\
Q2 & $N=10$, 30k & 0.206 & 0.116 & 0.565 & jSO-derived \\
Q2 & $N=12$, 10k & 0.265 & 0.201 & 0.759 & iL-SHADE \\
Q2 & $N=12$, 30k & 0.231 & 0.121 & 0.522 & jSO-derived \\
Q3 & $N=8$, 10k & 0.767 & 0.766 & 0.999 & SPSA \\
Q3 & $N=8$, 30k & 0.766 & 0.766 & 1.000 & SPSA \\
Q3 & $N=10$, 10k & 1.367 & 1.426 & 1.043 & SPSA \\
Q3 & $N=10$, 30k & 1.367 & 0.895 & 0.655 & jSO-derived \\
Q3 & $N=12$, 10k & 1.902 & 1.902 & 1.000 & SPSA \\
Q3 & $N=12$, 30k & 1.902 & 0.999 & 0.525 & SPSA \\
\bottomrule
\end{tabular}
\end{table*}

\keyresult{Adaptive rotation has a model-dependent VQE effect. On Q2 it reduces the iSOMA median in all four exact-objective conditions by 24--48\% and is significantly better than baseline in three of four direct comparisons. Q1 shows no benefit, while Q3 improves only at the larger FE budget.}

\paragraph{High-noise results}
The 25-run noisy experiment produces a pronounced ordering reversal relative to the exact Q2/Q3 study. Under high noise ($M=128$ effective shots), iSOMA-AR has the best mean rank across the six model--budget conditions (1.83), followed by baseline iSOMA (2.17), iL-SHADE (3.50), jSO-derived (3.67), CMA-ES (4.50), and SPSA (5.33). iSOMA-AR is ranked first or second in every high-noise condition. On Q2 and Q3 the two SOMA variants occupy the top two positions at both budgets; on Q1, jSO-derived is best and iSOMA-AR is second.

The effect is largest on Q2. At 10,000 FEs the exact error of the noisy-selected endpoint is 0.593 for iSOMA and 0.707 for iSOMA-AR, compared with 1.42 for CMA-ES, 2.51 for iL-SHADE, 2.54 for jSO-derived, and 4.21 for SPSA. At 30,000 FEs the SOMA medians are 0.433 and 0.522, while the nearest non-SOMA method is jSO-derived at 0.979. iSOMA-AR is significantly better than each non-SOMA comparator at both Q2 budgets after Holm correction.

High-noise Q3 shows the same family-level pattern. At 10,000 FEs, iSOMA and iSOMA-AR obtain 2.318 and 2.425, versus 3.239 for CMA-ES, 3.858 for iL-SHADE, 5.485 for SPSA, and 6.413 for jSO-derived. At 30,000 FEs, iSOMA-AR gives the smallest median, 2.177, closely followed by iSOMA at 2.221. No direct high-noise AR--baseline comparison is significant, pointing to the shared SOMA migration/acceptance mechanism as the more likely source of the observed robustness.

\paragraph{Low-noise results}
At the lower noise level ($M=8192$), the overall mean-rank advantage is smaller: iSOMA and iSOMA-AR tie at 3.00, followed by SPSA (3.50), iL-SHADE (3.67), CMA-ES (3.83), and jSO-derived (4.00). The result is strongly model dependent. On Q1, iSOMA is best and iSOMA-AR second at both budgets. On Q2, iSOMA-AR is best at 10,000 FEs, but SPSA, jSO-derived, and iL-SHADE overtake both SOMA variants at 30,000 FEs. On Q3, SPSA is clearly strongest at both budgets. No direct iSOMA-AR versus iSOMA comparison is significant after Holm correction at either noise level.

The offline oracle diagnostic separates search quality from noisy endpoint selection. Under high noise, the median condition-level selection penalty is 0.313 for iSOMA and 0.381 for iSOMA-AR, lower than CMA-ES (0.410), iL-SHADE (0.583), jSO-derived (0.620), and SPSA (0.707). The SOMA variants therefore combine strong search performance with comparatively small selection loss.

\paragraph{All-zero PRT-mask control.}
Forcing every PRT mask to activate at least one coordinate substantially reduced unchanged-point evaluations, but none of the eight paired Q1/Q2 endpoint comparisons was significant (all Wilcoxon $p\geq0.252$). All-zero masks therefore waste evaluations without materially changing the high-noise result. Full rates, paired ratios, and tests are reported in Appendix~\ref{app:prt_ablation}.

\begin{table*}[htpb]
\centering
\scriptsize
\caption{Sampling-noise robustness at $N=12$ over 25 runs. Entries are the median \emph{exact} errors of the parameter vector selected using the noisy objective. Bold marks the smallest median in each condition. Low and high noise correspond to 8192 and 128 effective shots, respectively.}
\label{tab:vqe_noise}
\begin{tabular}{@{}lllrrrrrr@{}}
\toprule
Noise & Model & Budget & iSOMA & iSOMA-AR & jSO-derived & iL-SHADE & CMA-ES & SPSA \\
\midrule
Low & Q1 & 10k & \textbf{0.066} & 0.087 & 0.115 & 0.177 & 1.712 & 2.022 \\
Low & Q1 & 30k & \textbf{0.062} & 0.080 & 0.083 & 0.158 & 1.755 & 2.017 \\
\addlinespace
Low & Q2 & 10k & 0.317 & \textbf{0.308} & 0.374 & 0.446 & 0.342 & 0.587 \\
Low & Q2 & 30k & 0.319 & 0.316 & 0.260 & 0.272 & 0.322 & \textbf{0.203} \\
\addlinespace
Low & Q3 & 10k & 1.928 & 1.927 & 3.689 & 1.555 & 1.382 & \textbf{0.356} \\
Low & Q3 & 30k & 1.907 & 1.910 & 1.960 & 1.244 & 1.216 & \textbf{0.166} \\
\midrule
High & Q1 & 10k & 1.037 & 0.958 & \textbf{0.686} & 1.003 & 1.952 & 1.717 \\
High & Q1 & 30k & 0.719 & 0.649 & \textbf{0.620} & 0.682 & 1.577 & 1.071 \\
\addlinespace
High & Q2 & 10k & \textbf{0.593} & 0.707 & 2.541 & 2.514 & 1.422 & 4.210 \\
High & Q2 & 30k & \textbf{0.433} & 0.522 & 0.979 & 1.610 & 1.662 & 2.290 \\
\addlinespace
High & Q3 & 10k & \textbf{2.318} & 2.425 & 6.413 & 3.858 & 3.239 & 5.485 \\
High & Q3 & 30k & 2.221 & \textbf{2.177} & 4.474 & 2.549 & 2.670 & 3.829 \\
\bottomrule
\end{tabular}
\end{table*}

\begin{figure*}[htpb]
    \centering
    \includegraphics[width=\textwidth]{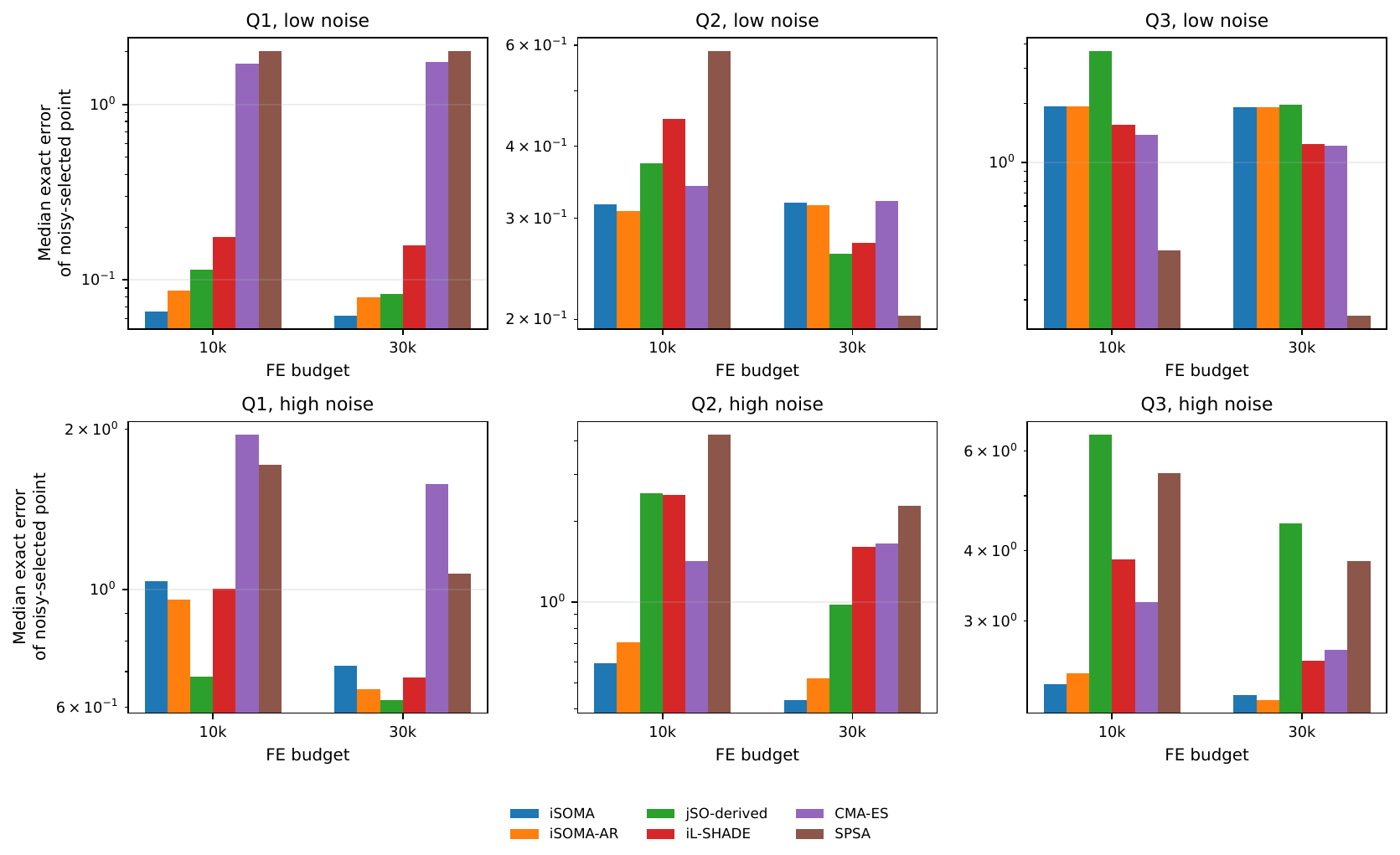}
    \caption{Median exact error of the parameter vector selected using the noisy objective in the full 25-run $N=12$ robustness experiment. Low and high noise correspond to 8192 and 128 effective shots, respectively.}
    \label{fig:vqe_noise_final}
\end{figure*}

\begin{figure*}[htpb]
    \centering
    \includegraphics[width=\textwidth]{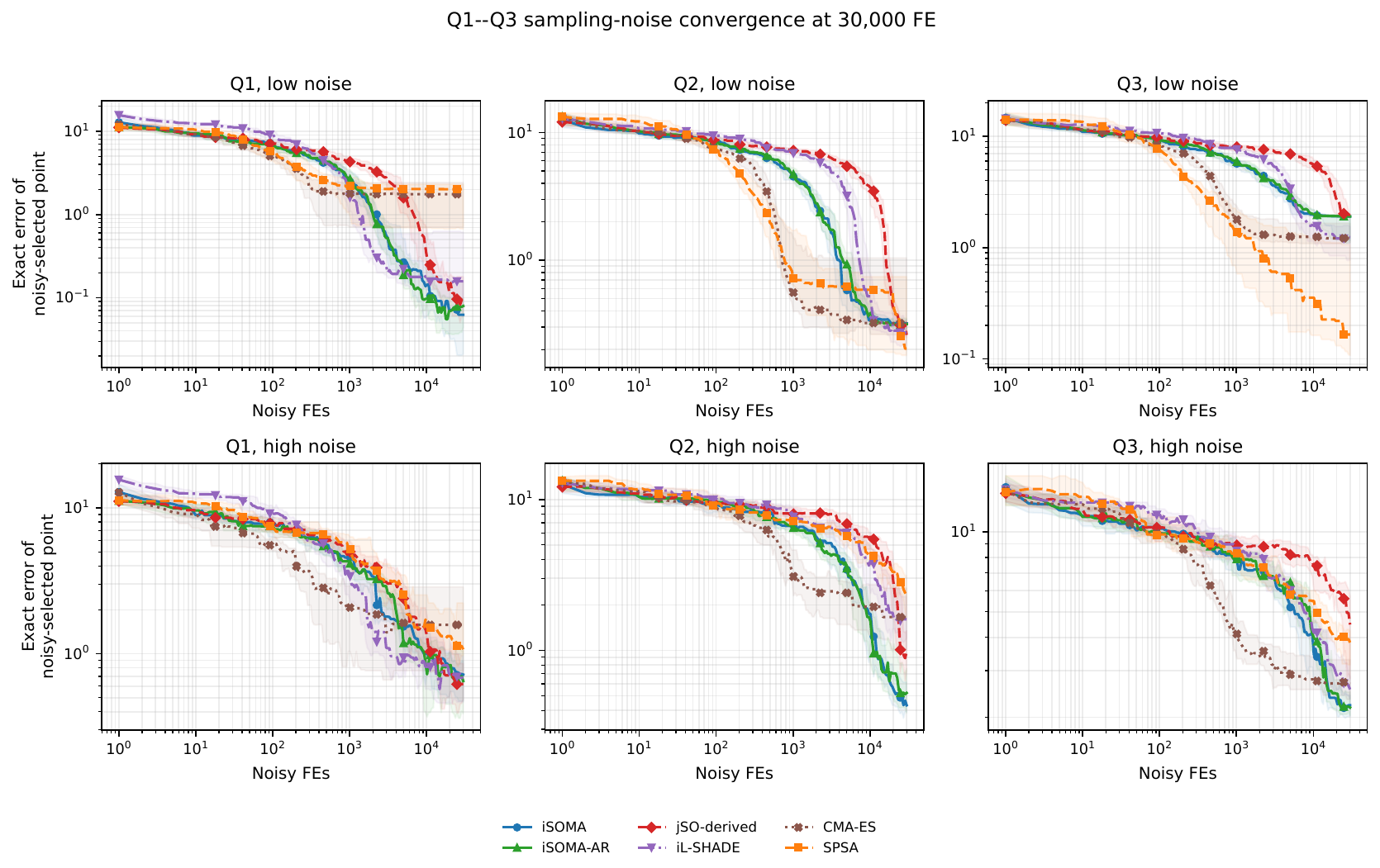}
    \caption{Sampling-noise convergence at $N=12$ and 30,000 FEs. Curves show the median exact error of the parameter vector selected by the noisy objective, with interquartile bands over 25 runs. Q1 and Q2 use physical ground-state error; Q3 uses optimization error relative to the saved best-known variational reference.}
    \label{fig:vqe_noise_convergence}
\end{figure*}

\section{Discussion}

\subsection{Interpretation across benchmark families}

Adaptive rotation is most useful when iSOMA's coordinate mask becomes a geometric limitation. On BBOB, iSOMA-AR improves the baseline in 507 of 720 matched conditions, increases fixed-target coverage, and shows its largest gains in the conditioned groups. Q2 follows the same pattern, with 24--48\% median reductions in all four exact conditions and three significant direct comparisons after Holm correction. Because Q1 and Q2 also differ in ansatz and parameter count, the synthetic ablation gives cleaner mechanistic evidence: on a rotated ill-conditioned Ellipsoid, learned AR significantly outperforms both baseline masking and a random-basis control using the same gate. Q1 shows no AR benefit, while Q3 improves only at the larger FE budget.

The combined evidence supports a conditional design principle: rotated masking helps when successful migrations reveal persistent off-axis structure. The gate limits unnecessary rotation, but Q1 shows that even conservative basis adaptation can hinder a proposal mechanism already aligned with the problem.

Important gaps remain. CMA-ES is markedly stronger on smooth anisotropic and ridge-like BBOB functions, while iL-SHADE is stronger on several structured-multimodal and rugged functions. Rotating the mask supplies less adaptation than a full covariance search distribution or modern DE scale and population control. On CEC 2011, iSOMA-AR ranks ahead of iL-SHADE overall and performs well on the power-system subset, but its aggregate improvement over baseline iSOMA is not significant.

\subsection{Noise robustness of the SOMA migration mechanism}

The full 25-run VQE noise experiment reveals a second, largely independent property of the SOMA family. Under high noise, iSOMA-AR has the best mean rank and is first or second in all six model--budget conditions, while baseline iSOMA ranks second overall. On Q2 and Q3 the two SOMA variants occupy the top two positions at both budgets. This is a substantial reversal relative to the exact Q3 ordering and to several exact Q2 conditions.

None of the 12 noisy AR--baseline comparisons is significant after Holm correction, so the high-noise advantage is better attributed to the shared SOMA migration/acceptance process than to adaptive rotation. iSOMA and iSOMA-AR also have the two smallest median selection penalties among the six methods. The all-zero-PRT control reaches the same conclusion from another angle: repairing zero masks sharply reduces unchanged-point proposals but does not significantly change any of the eight paired Q1/Q2 endpoints. Reverse-path evaluation, first-improvement acceptance, and overshoot are plausible contributors to the robustness, but separating their effects requires a dedicated ablation.

At lower noise, the two SOMA methods tie for the best mean rank, but SPSA dominates Q3 and the DE/CMA methods become competitive on Q2. The SOMA advantage therefore emerges mainly in the stronger-noise regime, where fine-grained ranking and adaptation are less reliable.

\subsection{Comparator behavior and computational cost}

The jSO-derived comparator is also notable: it is statistically indistinguishable from CMA-ES in aggregate BBOB performance and significantly stronger on CEC 2011 under the present protocol. The canonical-jSO rerun clarifies the implementation dependence. Canonical jSO remains competitive, but the derived implementation is stronger on the aggregate BBOB sensitivity comparison; both are reported separately.

The computational overhead of iSOMA-AR remains modest. On BBOB, its median cost per 1000 objective evaluations is 1.12 times the baseline, versus 1.54 for jSO-derived, 1.84 for iL-SHADE, and 3.72 for CMA-ES. On CEC 2011 the adaptive basis is relatively more expensive (1.75 times baseline) because several problems have much larger dimension, but it remains much cheaper per FE than CMA-ES. For expensive simulation or quantum-circuit objectives, this optimizer-side overhead should be comparatively less important than the evaluation cost itself.

\subsection{Limitations and next experiments}

Several limitations remain. The CEC 2011 experiment uses one 50,000-evaluation budget and ten runs rather than the original competition's larger multi-budget protocol, and four Minion CEC 2011 functions (F3, F4, F21, F22) require independent validation against the reference implementation. The CEC normalized-target ECDF is diagnostic rather than an official benchmark metric.

The VQE benchmark uses 25 independent runs per condition, but its additive Gaussian noise model omits state-dependent Pauli variances, measurement grouping, shot allocation, readout error, and hardware drift. An explicit finite-shot estimator on Q2 and Q3 is the natural next test. The AR ablation is limited to two controlled landscape pairs and six BBOB functions, and the zero-PRT experiment isolates only one implementation detail; path order, first-improvement acceptance, overshoot, and step/path-length adaptation remain open ablations. Canonical jSO was rerun only for BBOB and CEC 2011; VQE uses jSO-derived.

A natural extension is to adaptively choose among coordinate masking, rotated masking, and step/path-length settings using recent successful moves and evaluation efficiency.

\section{Conclusion}
iSOMA-AR provides a targeted geometric correction to iSOMA. It improves the baseline robustly on BBOB and gives its clearest VQE gain on Q2, while Q1 shows no benefit. The learned-versus-random control confirms the value of the learned orientation on a rotated ill-conditioned landscape, and moderate gate-parameter changes preserve the qualitative result. Under strong sampling noise, both SOMA variants are highly competitive, but AR is not significantly better than baseline and zero-mask repair does not alter the outcome. The canonical-jSO rerun further separates the strong custom jSO-derived results from canonical jSO. Overall, adaptive rotation improves deterministic handling of coupled geometry, while the observed high-noise robustness is primarily a property of the underlying SOMA migration mechanism.

\section*{CRediT authorship contribution statement}
\textbf{Vojtěch Novák}: Software, Investigation, Writing – original draft. \textbf{Ivan Zelinka}: Supervision, Validation, Reviewing.

\section*{Acknowledgements}
This project has received funding from the Research Council of Lithuania (LMTLT), agreement No.: S-ITP-25-7 This research was also supported by research grants SGS No. SP2026/063 of VSB-Technical University of Ostrava, Czech Republic.

\section*{Declaration of Competing Interest}
The authors declare that they have no known competing financial interests or personal relationships that could have appeared to influence the work reported in this paper.

\section*{Data availability}
All simulations were conducted using a locally developed code, which is openly available for further examination and reproduction of results. The iSOMA-AR code and scripts for reproducing the results can be accessed at \url{https://github.com/VojtechNovak/iSOMA-AR}. Further details are available through the corresponding author.

\appendix

\section{Optimizer configurations and implementation details}
\label{app:optimizers}

Table~\ref{tab:optimizer_configurations} records the optimizer implementations used in the reported experiments. The BBOB and CEC 2011 drivers share the same optimizer functions; the VQE experiment preserves its fixed legacy wrappers where noted.

\begin{table}[htpb]
\centering
\scriptsize
\setlength{\tabcolsep}{3pt}
\renewcommand{\arraystretch}{0.90}
\caption{Optimizer implementations and numerical settings. $B$ denotes the FE budget and $D$ the decision-space dimension.}
\label{tab:optimizer_configurations}
\begin{tabularx}{\linewidth}{@{}
  >{\raggedright\arraybackslash\hsize=0.60\hsize}X
  >{\raggedright\arraybackslash\hsize=0.95\hsize}X
  >{\raggedright\arraybackslash\hsize=1.45\hsize}X
  >{\raggedright\arraybackslash\hsize=1.00\hsize}X
@{}}
\toprule
Method & Role / implementation & Settings used & Notes \\
\midrule
iSOMA &
Baseline SOMA used in all three benchmark families. &
Pop $50$; $N_{\mathrm{jump}}=10$; Step $=0.3$; $m=10$, $n=5$, $k=15$; masks regenerated every proposal; $\mathrm{PRT} = 0.1+0.9\,N_{\mathrm{FE}}/B$; path $t=3.0,2.7,\ldots,0.3$; first non-worsening proposal accepted; box clipping; 10\% pop reinitialization after $>50\,\mathrm{PopSize}$ failed attempts without global best improvement. &
FE budget is effective stopping rule. BBOB/CEC driver uses nonbinding migration cap of $10^5$; VQE wrapper uses $10^4$. \\
\addlinespace[3pt]
iSOMA-AR &
Proposed variant; identical baseline selection, path, repair, and acceptance with occasional rotated masking. &
All iSOMA settings above plus $C_s(0)=I$, success update scale $0.12$ (cap $0.35$), eigenspace update every 5 accepted improving steps, ridge $10^{-4}I$, min.\ successful-step count $\max(6,\lceil1.5D\rceil)$, off-axis threshold $0.18$, max rotation probability $0.45$. &
No extra FEs spent on basis learning. Covariance memory and rotation gate reset at baseline stagnation restart. \\
\addlinespace[3pt]
jSO-derived &
Fixed custom DE used throughout BBOB/CEC and VQE comparison. &
$N_0=\max(30,\lfloor25\sqrt{D}\log_{10}(\max(D,2))\rfloor)$, linear reduction to $N_{\min}=4$; $H=5$; $M_F=0.5$, $M_{CR}=0.8$; Cauchy $F$ (scale $0.1$), Gaussian $CR$ ($\sigma=0.1$); $p$ decreases linearly $0.25 \to 0.125$; external archive with pop capacity; binomial crossover; clipping repair; improvement-weighted memory. &
Trial mutation: $x_i+F(x_{p\mathrm{best}}-x_i)+F(x_{r1}-x_{r2})$. Reported as \emph{jSO-derived}; specific differences are listed below. \\
\addlinespace[3pt]
jSO-canonical &
Post-hoc BBOB/CEC implementation-sensitivity comparator following the published jSO/iL-SHADE mechanisms. &
$N_0=\max(4,\mathrm{round}(25\ln(\max(D,2))\sqrt D))$, linear reduction to $N_{\min}=4$; $H=5$; $M_F=0.3$, $M_{CR}=0.8$ with the special final $(0.9,0.9)$ memory entry; $p:0.25\to0.125$; early $CR$ lower bounds and $F$ cap; weighted current-to-$p$best mutation; external archive; midpoint-to-parent bound repair. &
Uses the paper-prose $p$ schedule. Run on the complete BBOB and CEC 2011 grids as a sensitivity comparator; VQE uses jSO-derived. \\
\addlinespace[3pt]
iL-SHADE &
Standardized adaptive-DE reference through PyADE. &
\texttt{pyade.ilshade.get\_default\_params(D)} with pop size $4D$, explicit box bounds, supplied replicate seed, and \texttt{max\_evals}$=B$. All other settings are package defaults. &
No internal fallback results are included in reported experiments. \\
\addlinespace[3pt]
CMA-ES &
Covariance-adaptation reference. &
BBOB/CEC: \texttt{pycma}, uniform $x_0$, $\sigma_0=0.15 \times \text{mean box width}$, \texttt{maxfevals}$=B$, box bounds, supplied seed. VQE: fixed wrapper with $\lambda=4+\lfloor3\ln D\rfloor$, $\sigma_0=1$, \texttt{tolx}=\texttt{tolfun}=$10^{-10}$, generation cap $\lfloor B/\lambda\rfloor$. &
Objective-level FE guard enforces VQE budget if internal termination occurs later. \\
\addlinespace[3pt]
SPSA &
Stochastic-gradient reference for VQE exact archive and noise tests. &
Uniform $x_0\in[-\pi,\pi]^D$; $a=0.30$, $c=0.10$, $A=100$, $\alpha=0.602$, $\gamma=0.101$; one Rademacher perturbation/iter (2 FEs); parameters wrapped periodically to $[-\pi,\pi]^D$. &
Gradient estimate is explicitly designed for noisy function values \citep{spall1992spsa}. \\
\addlinespace[3pt]
BFGS &
Auxiliary exact-VQE comparator only. &
SciPy BFGS with numerical finite-difference gradient, uniform $x_0$, and \texttt{gtol}$=10^{-8}$. Q1 uses \texttt{maxiter}$=1000$; Q2/Q3 use $\max(1,\lfloor B/(D+1)\rfloor)$. &
Excluded from the noisy VQE panel because finite-difference BFGS is unsuitable as the target stochastic baseline. \\
\addlinespace[3pt]
SciPy-DE &
Auxiliary exact-VQE comparator only. &
Pop $2D$ (\texttt{popsize}=2); \texttt{currenttobest1bin}; mutation dither $(0.4,0.9)$; recombination $0.8$; \texttt{tol}=0; no polishing; deferred updating; strict FE guard. &
Retained in exact archive; excluded from BBOB/CEC headline comparison. \\
\addlinespace[3pt]
L-SRTDE &
Auxiliary exact-VQE adaptive-DE comparator only. &
Initial pop $20D$, linear reduction to $4$; memory $H=5$ for $CR$; adaptive $\mu_F=0.4+0.25\tanh(5\,\text{success rate})$ with $\sigma_F=0.02$; adaptive $p$-best size, rank-weighted elite base; random reinitialization on bound violation. &
Imported from supplied implementation; run under same objective-level FE cap. \\
\bottomrule
\end{tabularx}
\end{table}

\paragraph{Why the comparator is labelled jSO-derived.}
Published jSO uses the weighted current-to-$p$best mutation
\begin{equation}
    v_i
    =
    x_i
    +
    F_w(x_{p\mathrm{best}}-x_i)
    +
    F(x_{r1}-x_{r2}),
\end{equation}
with $F_w=0.7F$, $0.8F$, and $1.2F$ over successive budget phases \citep{brest2017jso}. The published description also specifies $N_0=25\log(D)\sqrt D$, $M_F=0.3$, one fixed $0.9$ memory entry inherited from iL-SHADE, early-stage lower bounds on $CR$, an early cap on $F$, and its own bound-repair procedure. The implementation used here explicitly uses $\log_{10}D$ with a floor of 30 individuals, initializes every $M_F$ entry to $0.5$, applies the same $F$ to the $p$best and differential terms, does not impose the published stage-dependent $F/CR$ restrictions, and clips out-of-bound trial coordinates. It retains the jSO/iL-SHADE ideas of success-history parameter adaptation, dynamic $p$, an external archive, and linear population-size reduction. These differences are substantial, so the implementation is reported consistently as \emph{jSO-derived}.

\paragraph{Canonical-jSO sensitivity implementation.}
The post-hoc comparator restores the principal published jSO mechanisms: natural-log population initialization, $M_F=0.3$, the special final $(0.9,0.9)$ memory entry, stage-dependent restrictions on $F$ and $CR$, the weighted current-to-$p$best term, an external archive, linear population-size reduction, and midpoint-to-parent bound repair. The jSO paper is internally ambiguous about the printed $p$ schedule; the reported sensitivity run follows the prose convention in which $p$ decreases from 0.25 to 0.125. The rerun is reported as \emph{jSO-canonical} and used as a post-hoc sensitivity comparator.

\subsection{Runtime overhead}
\label{app:overhead}
The relatively low overhead of iSOMA-AR reflects the sparse form of its adaptation. Most proposals retain the \(O(D)\) baseline masking operation; dense \(O(D^2)\) rotations are gated and probabilistic, while the eigendecomposition of the success matrix is performed only after every five improving updates. In contrast, the DE variants perform mutation, crossover, population/archive management, and parameter adaptation for each trial, while CMA-ES maintains and updates a full covariance-based search distribution. The larger relative AR overhead on CEC 2011 is consistent with its higher problem dimensions. For expensive simulation or quantum objectives, these optimizer-side differences are expected to be small relative to the cost of an objective evaluation.

\begin{table}[htpb]
\centering
\caption{Median computational overhead normalized per 1000 objective evaluations. Wall-clock values are implementation- and hardware-dependent and are included only as relative overhead indicators.}
\label{tab:runtime}
\small
\begin{tabular}{@{}lrrrr@{}}
\toprule
Algorithm & BBOB s/1000 FE & vs base & CEC s/1000 FE & vs base \\
\midrule
iSOMA & 0.0182 & 1.00$\times$ & 0.0110 & 1.00$\times$ \\
iSOMA-AR & 0.0204 & 1.12$\times$ & 0.0193 & 1.75$\times$ \\
jSO-derived & 0.0279 & 1.54$\times$ & 0.0202 & 1.83$\times$ \\
jSO-canonical$^{\dagger}$ & 0.0304 & 1.67$\times$ & 0.0253 & 2.30$\times$ \\
iL-SHADE & 0.0334 & 1.84$\times$ & 0.0224 & 2.03$\times$ \\
CMA-ES & 0.0676 & 3.72$\times$ & 0.0834 & 7.55$\times$ \\
\bottomrule
\end{tabular}
\begin{flushleft}
\scriptsize $^{\dagger}$Post-hoc sensitivity run; wall-clock values were measured separately from the primary panel and are included only as an implementation-overhead indication.
\end{flushleft}
\end{table}

\section{VQE instance and reference construction}
\label{app:vqe_details}

\paragraph{Ising-backbone selection.}
For each required Ising size, the degree-three ring-plus-matching graph is fixed first. A master generator with seed 1234 produces 50 candidate disorder seeds. Each candidate draws $J^z_{ij}\in\{-1,+1\}$ and $h_i\sim U[-0.2,0.2]$, after which all $2^N$ diagonal energies are enumerated and the number of one-spin-flip local minima is counted exactly. The candidate with the largest count is retained. These arrays, their ground energy, and the instance seed are stored and reused for every optimizer. Q2 never generates a replacement disorder realization; it loads the matched $J^z,h$ arrays and adds only the transverse-field term.

\paragraph{Heisenberg disorder.}
For Q3 at $N=10,12$, the $J^z,h$ backbone is the same stored Ising realization used by the corresponding Q1/Q2 construction. The $J^x$ and $J^y$ couplings are independent $\pm1$ draws taken at fixed RNG positions from seed 1234. At $N=8$, where no paired Ising benchmark condition is needed, the complete $J^z,J^x,J^y,h$ realization is fixed by the same seed-1234 construction. This policy preserves the Q3-specific transverse couplings while allowing the $N=10,12$ Hamiltonians to share the same classical frustrated backbone as Q2.

\paragraph{Q3 variational reference.}
The shallow $2N$-parameter circuit does not in general contain the exact Heisenberg ground state. For $N=10$ and $N=12$, the best-known variational reference is constructed from 20 independent SciPy differential-evolution searches using \texttt{best1bin}, \texttt{popsize}=30, 200 generations, and tolerance $10^{-10}$, each followed by BFGS polishing with \texttt{gtol}=$10^{-10}$ and at most 5000 iterations. The lowest polished energy is retained. For $N=8$, a fixed reference energy is retained after consistency checks against archived optimizer endpoints. During the benchmark, any endpoint improving the stored Q3 reference by more than $10^{-12}$ is allowed to replace it and all Q3 optimization errors are then recomputed relative to the final value. The iSOMA-AR extension did not change any final Q3 reference.

\section{Adaptive-rotation mechanism ablation}
\label{app:ar_ablation}

The revision ablation is targeted to the proposed mechanism. The controlled landscapes distinguish whether improvement comes from the learned orientation itself or merely from injecting a rotated mask. The random-basis control uses the same gate score and rotation probability as learned AR, but replaces the learned eigenbasis by a fixed random orthogonal basis for each run. Threshold and rotation-cap variants change one gate parameter at a time.

\begin{table*}[htpb]
\centering
\small
\caption{Controlled adaptive-rotation mechanism ablation at $D=10$ and 20,000 FEs (25 paired runs). W/T/L and the median error ratio compare the listed control with default learned AR; ratios above one favor learned AR.}
\label{tab:ar_mechanism_control}
\begin{tabular}{@{}llrrrr@{}}
\toprule
Problem & Control / default AR & W/T/L & Median ratio & Wilcoxon $p$ & AR rotation frac. \\
\midrule
Ellipsoid (axis-aligned) & random basis / AR default & 10/0/15 & 1.41 & 0.692 & 7.4\% \\
Ellipsoid (axis-aligned) & iSOMA / AR default & 7/0/18 & 12.74 & 0.0451 & 7.4\% \\
Ellipsoid (rotated) & random basis / AR default & 3/0/22 & 3.16 & \num{1.20e-04} & 22.5\% \\
Ellipsoid (rotated) & iSOMA / AR default & 6/0/19 & 4.33 & \num{1.20e-04} & 22.5\% \\
Rastrigin (axis-aligned) & random basis / AR default & 11/0/14 & 1.00 & 0.874 & 12.4\% \\
Rastrigin (axis-aligned) & iSOMA / AR default & 14/0/11 & 0.90 & 0.491 & 12.4\% \\
Rastrigin (rotated) & random basis / AR default & 13/0/12 & 0.86 & 0.916 & 15.9\% \\
Rastrigin (rotated) & iSOMA / AR default & 12/0/13 & 1.14 & 0.916 & 15.9\% \\
\bottomrule
\end{tabular}
\end{table*}

\begin{figure*}[htbp]
    \centering
    \includegraphics[width=0.90\textwidth]{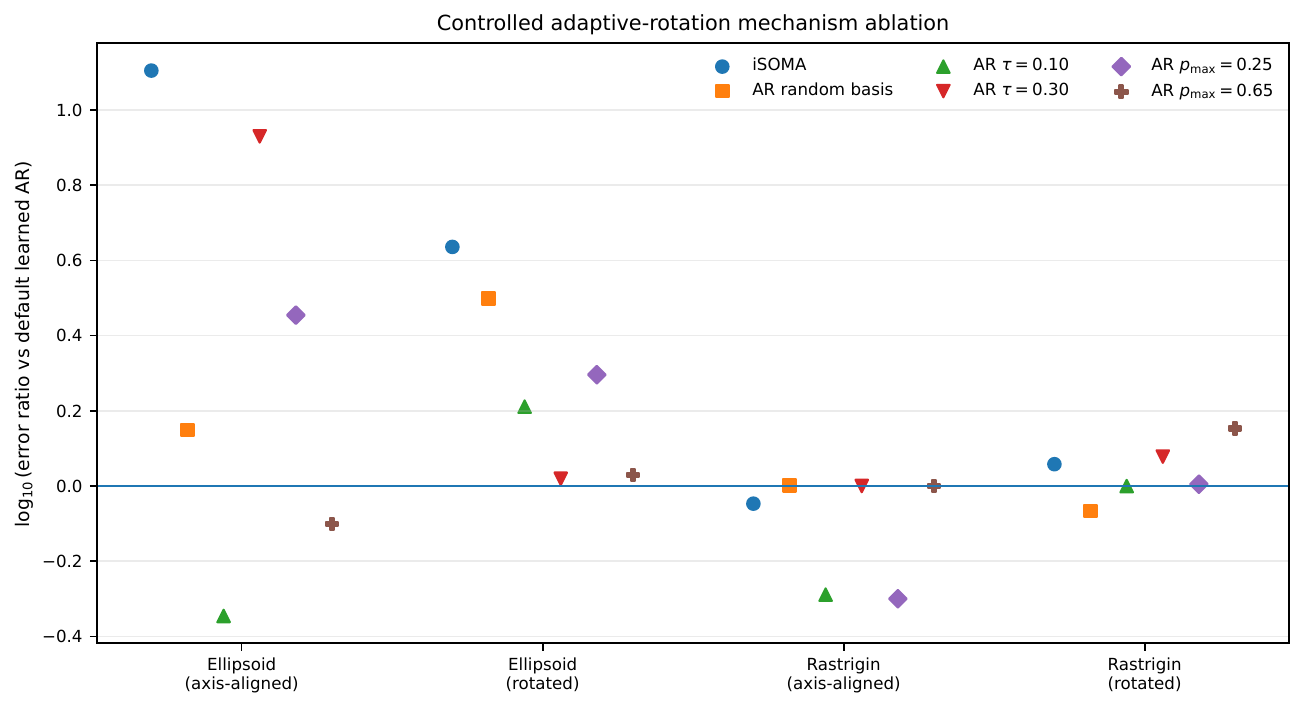}
    \caption{Controlled AR ablation. Points show the median paired error ratio of each control relative to default learned AR. Values above zero on the log scale favor default learned AR.}
    \label{fig:ar_ablation_controlled}
\end{figure*}

The rotated Ellipsoid gives the clearest mechanism control: learned AR is significantly better than both baseline iSOMA and the random-basis variant. The corresponding random-basis comparison is not significant on the axis-aligned Ellipsoid, and neither Rastrigin pair shows a significant learned-orientation effect. The sensitivity sweep is broadly stable; the largest degradation occurs when the maximum rotation probability is reduced to 0.25 on the rotated Ellipsoid.

\begin{table}[htpb]
\centering
\caption{One-factor sensitivity of the adaptive-rotation gate. Each variant is paired against the default setting $\tau=0.18$, $p_{\max}=0.45$.}
\label{tab:ar_parameter_sensitivity}
\begin{tabular}{@{}llrrr@{}}
\toprule
Problem & Variant & W/T/L & Median ratio & Wilcoxon $p$ \\
\midrule
Ellipsoid (axis-aligned) & AR $p_{\max}=0.25$ & 11/0/14 & 2.85 & 0.23 \\
Ellipsoid (axis-aligned) & AR $p_{\max}=0.65$ & 13/0/12 & 0.79 & 0.916 \\
Ellipsoid (axis-aligned) & AR $\tau=0.10$      & 15/0/10 & 0.45 & 0.474 \\
Ellipsoid (axis-aligned) & AR $\tau=0.30$      & 9/0/16  & 8.51 & 0.21 \\
\addlinespace
Ellipsoid (rotated)      & AR $p_{\max}=0.25$ & 6/0/19  & 1.98 & 0.0125 \\
Ellipsoid (rotated)      & AR $p_{\max}=0.65$ & 12/0/13 & 1.07 & 0.23 \\
Ellipsoid (rotated)      & AR $\tau=0.10$      & 7/0/18  & 1.63 & 0.191 \\
Ellipsoid (rotated)      & AR $\tau=0.30$      & 11/0/14 & 1.05 & 0.771 \\
\addlinespace
Rastrigin (axis-aligned) & AR $p_{\max}=0.25$ & 14/0/11 & 0.50 & 0.287 \\
Rastrigin (axis-aligned) & AR $p_{\max}=0.65$ & 14/0/11 & 1.00 & 0.458 \\
Rastrigin (axis-aligned) & AR $\tau=0.10$      & 15/0/10 & 0.51 & 0.23 \\
Rastrigin (axis-aligned) & AR $\tau=0.30$      & 13/0/12 & 1.00 & 0.634 \\
\addlinespace
Rastrigin (rotated)      & AR $p_{\max}=0.25$ & 10/0/15 & 1.01 & 0.156 \\
Rastrigin (rotated)      & AR $p_{\max}=0.65$ & 8/0/17  & 1.42 & 0.252 \\
Rastrigin (rotated)      & AR $\tau=0.10$      & 14/0/11 & 1.00 & 0.751 \\
Rastrigin (rotated)      & AR $\tau=0.30$      & 11/0/14 & 1.20 & 0.426 \\
\bottomrule
\end{tabular}
\end{table}

The selected BBOB check uses matched natural/rotated or structurally related functions. Learned AR is significantly better than both baseline and the random-basis control on rotated Rosenbrock $f_9$ and rotated Ellipsoid $f_{10}$ after pooling the two tested instances. On $f_2$, the random-basis control is competitive, showing that the learned gate is not uniformly advantageous.

\begin{table*}[htpb]
\centering
\small
\caption{Selected BBOB corroboration of the AR mechanism at $D=10$, instances 1--2, 2000 FE/D (20 paired runs per function after pooling instances). Ratios above one favor default learned AR.}
\label{tab:ar_bbob_ablation}
\begin{tabular}{@{}lrrrrrr@{}}
\toprule
Function & base W/T/L & base/AR & $p$ & random W/T/L & random/AR & $p$ \\
\midrule
$f_{2}$ & 6/0/14 & 6.88 & 0.0973 & 14/0/6 & 0.05 & 0.0637 \\
$f_{3}$ & 10/0/10 & 1.00 & 0.784 & 10/0/10 & 1.00 & 0.898 \\
$f_{8}$ & 4/0/16 & 3.53 & 0.00639 & 8/0/12 & 1.79 & 0.245 \\
$f_{9}$ & 4/0/16 & 1.41 & 0.00365 & 3/0/17 & 1.67 & \num{7.08e-04} \\
$f_{10}$ & 4/0/16 & 1.94 & 0.00315 & 4/0/16 & 3.01 & 0.00315 \\
$f_{15}$ & 6/0/14 & 1.37 & 0.114 & 11/0/9 & 0.84 & 0.596 \\
\bottomrule
\end{tabular}
\end{table*}

\begin{figure*}[htbp]
    \centering
    \includegraphics[width=0.72\textwidth]{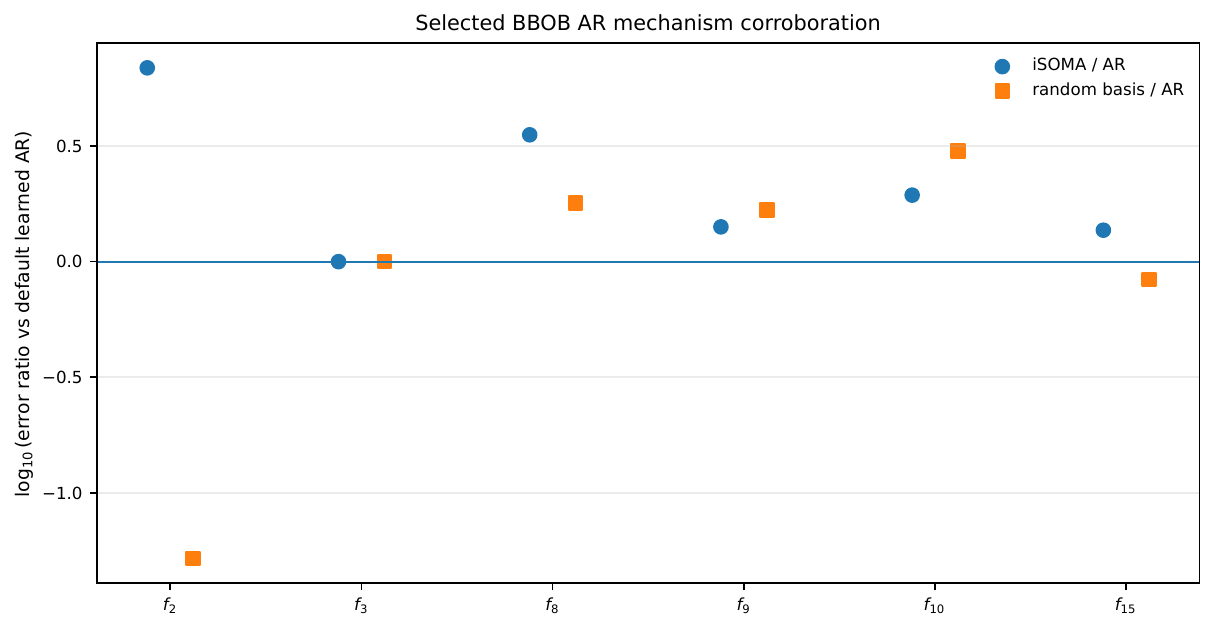}
    \caption{Selected BBOB mechanism corroboration. Ratios compare baseline iSOMA or random-basis AR against default learned AR; positive log-ratios favor learned AR.}
    \label{fig:ar_ablation_bbob}
\end{figure*}

\section{All-zero PRT-mask control}
\label{app:prt_ablation}

The baseline mask samples coordinates independently and can therefore be all zero. This evaluates the unchanged point and, under noise, obtains another noisy observation at the same location. The paired control activates one uniformly random coordinate whenever this occurs, using a separate repair RNG. The paired variants otherwise share the optimizer seed and FE-indexed Gaussian-noise stream.

\begin{table*}[htpb]
\centering
\small
\caption{All-zero PRT-mask control under high effective-shot Gaussian noise at $N=12$ (25 paired runs). The nonzero variant repairs an all-zero Bernoulli mask by activating one uniformly random coordinate. Performance ratios below one favor the repair.}
\label{tab:prt_zero_mask}
\begin{tabular}{@{}llrrrrrr@{}}
\toprule
Model & Algorithm & FE & W/T/L & Error ratio & Wilcoxon $p$ & Same-point allowed & Same-point repaired \\
\midrule
Q1 & iSOMA & 10k & 14/0/11 & 0.90 & 0.833 & 2.34\% & 0.50\% \\
Q1 & iSOMA & 30k & 15/0/10 & 0.88 & 0.596 & 2.71\% & 0.92\% \\
Q2 & iSOMA & 10k & 10/0/15 & 1.18 & 0.396 & 0.49\% & 0.46\% \\
Q2 & iSOMA & 30k & 9/0/16 & 1.15 & 0.252 & 0.91\% & 0.40\% \\
Q1 & iSOMA-AR & 10k & 14/0/11 & 0.82 & 0.672 & 2.26\% & 0.37\% \\
Q1 & iSOMA-AR & 30k & 14/0/11 & 0.96 & 0.833 & 2.56\% & 0.53\% \\
Q2 & iSOMA-AR & 10k & 14/0/11 & 0.94 & 0.751 & 0.38\% & 0.15\% \\
Q2 & iSOMA-AR & 30k & 14/0/11 & 0.87 & 0.731 & 0.55\% & 0.20\% \\
\bottomrule
\end{tabular}
\end{table*}

\begin{figure*}[htbp]
    \centering
    \includegraphics[width=0.82\textwidth]{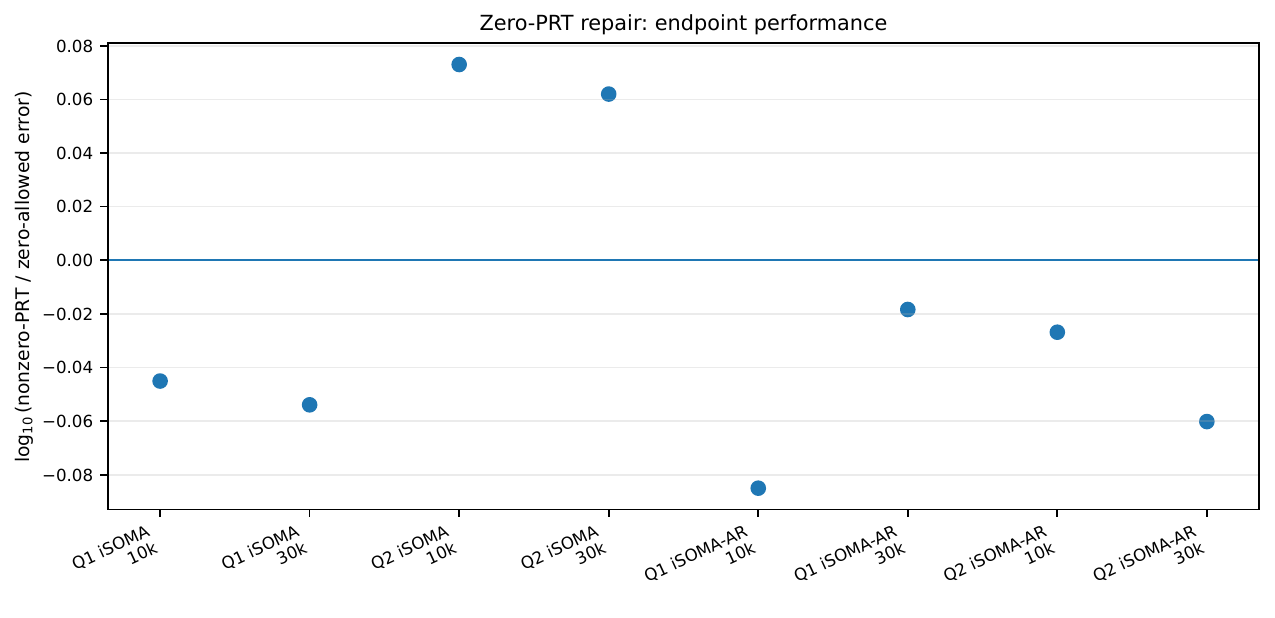}
    \caption{Effect of forcing nonzero PRT masks on noisy-selected endpoint quality. Values near zero indicate little change; none of the eight paired comparisons is statistically significant.}
    \label{fig:prt_zero_mask_performance}
\end{figure*}

\begin{figure*}[htbp]
    \centering
    \includegraphics[width=0.82\textwidth]{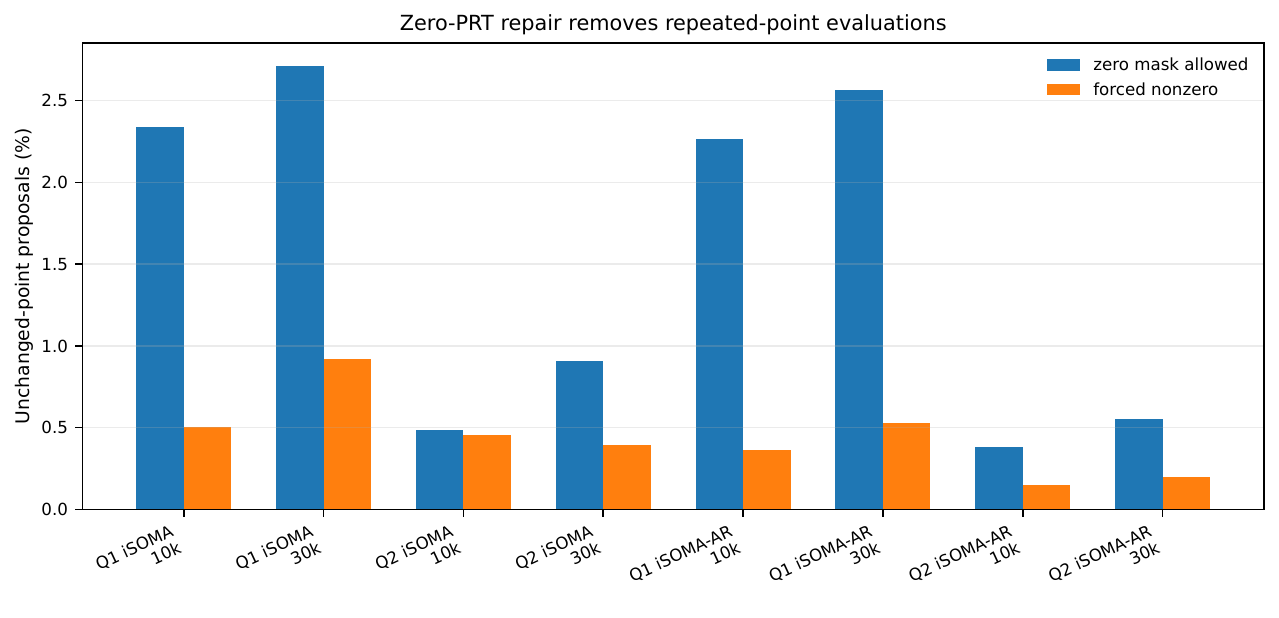}
    \caption{Fraction of proposals that re-evaluate an unchanged point when all-zero PRT masks are allowed or repaired. The repair substantially reduces repeated-point evaluations, especially for Q1, without a corresponding significant endpoint improvement.}
    \label{fig:prt_zero_mask_same_point}
\end{figure*}

\section{Canonical-jSO sensitivity analysis}
\label{app:jso_sensitivity}

Paper-faithful canonical jSO was rerun on the complete BBOB grid (7,200 runs) and CEC 2011 grid (220 runs) with the same budgets and seed labels as jSO-derived. The original five-method ranks remain the primary analysis. For CEC target metrics, fallback reference values are kept fixed to the original comparator panel.

\begin{table*}[htpb]
\centering
\small
\setlength{\tabcolsep}{4pt}
\caption{Post-hoc canonical-jSO sensitivity analysis. Ranks and Top-1 shares use the six-method sensitivity panels. The tight target is $\Delta f=10^{-8}$ for BBOB and the $10^{-6}$ relative-gap target for CEC 2011; tight runtime is FE/D for BBOB and a fraction of the 50,000-FE budget for CEC. Direct rows compare jSO-derived with jSO-canonical, and ratios below one favor jSO-derived.}
\label{tab:jso_sensitivity}
\begin{tabular}{@{}llrrrrr@{}}
\toprule
Study & Algorithm & Mean rank & Top-1 & All targets & Tight target & Tight runtime \\
\midrule
BBOB & jSO-derived   & 2.00 & 46.8\% & 65.4\% & 48.8\% & $1.08\times10^3$ \\
BBOB & jSO-canonical & 3.33 & 18.6\% & 57.8\% & 36.8\% & $1.52\times10^3$ \\
\multicolumn{7}{@{}l}{\emph{Direct derived/canonical:} 557/104/59 W/T/L; median error ratio 0.48; Wilcoxon $p=\num{6.32e-79}$.} \\
\addlinespace
CEC 2011 & jSO-derived   & 1.77 & 63.6\% & 62.9\% & 29.1\% & 0.504 \\
CEC 2011 & jSO-canonical & 2.73 & 27.3\% & 56.7\% & 23.6\% & 0.222 \\
\multicolumn{7}{@{}l}{\emph{Direct derived/canonical:} 15/2/5 W/T/L; median regret ratio 0.47; Wilcoxon $p=0.0853$.} \\
\bottomrule
\end{tabular}
\end{table*}

The VQE experiment uses jSO-derived only.

\section{Additional BBOB, CEC 2011 and VQE results}

\begin{table*}[htpb]
\centering
\caption{Diagnostic BBOB groups used to separate landscape mechanisms. These groups overlap by design.}
\label{tab:bbob_custom}
\scriptsize
\resizebox{\textwidth}{!}{%
\begin{tabular}{@{}lp{2.9cm}rrrr@{}}
\toprule
Group & Frequent winner & AR/base W/T/L & AR/base ratio & AR/iLS W/T/L & AR/iLS ratio \\
\midrule
C1 ill-conditioned / anisotropic & CMA-ES 137/210 & 186/2/22 & 0.49 & 60/2/148 & 2.72 \\
C2 valley / ridge path-following & CMA-ES 122/150 & 109/0/41 & 0.80 & 35/0/115 & 1.76 \\
C3 structured multimodality & jSO-derived 125/210 & 138/0/72 & 0.91 & 80/0/130 & 1.42 \\
C4 basin selection / weak structure & jSO-derived 60/120 & 72/3/45 & 0.99 & 65/3/52 & 0.92 \\
C5 plateau / nonsmooth / rugged & jSO-derived 48/120 & 83/0/37 & 0.86 & 24/0/96 & 1.67 \\
C6 asymmetry / deception & CMA-ES 42/90 & 63/0/27 & 0.92 & 39/0/51 & 1.37 \\
C7 boundary / extrapolation & jSO-derived 48/60 & 11/30/19 & 1.00 & 37/23/0 & 0.87 \\
\bottomrule
\end{tabular}
}
\end{table*}

\begin{figure*}[htpb]
    \centering
    \includegraphics[width=\textwidth]{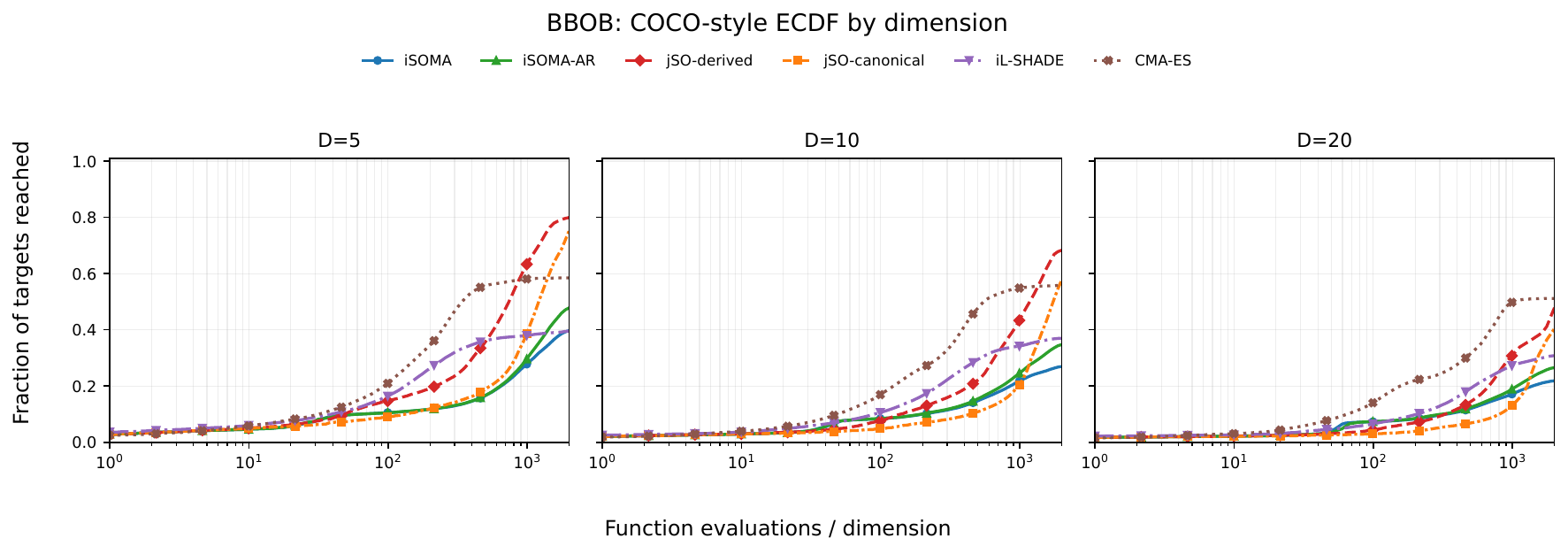}
    \caption{BBOB COCO-style fixed-target ECDF split by dimension.}
    \label{fig:bbob_dims}
\end{figure*}

\begin{table}[htpb]
\centering
\footnotesize 
\setlength{\tabcolsep}{3.5pt}
\renewcommand{\arraystretch}{0.92}
\caption{Function-level BBOB outcomes across 30 matched dimension--instance--budget conditions per function. AR/base and AR/iLS are W/T/L for iSOMA-AR against baseline iSOMA and iL-SHADE.}
\label{tab:bbob_functions}
\begin{tabularx}{\linewidth}{@{} r l >{\raggedright\arraybackslash}X >{\raggedright\arraybackslash}p{2.6cm} ccc @{}}
\toprule
$f$ & Function & Landscape feature & Frequent winner & AR/base & AR/iLS & jSO-derived/CMA \\
\midrule
1  & Sphere                     & smooth, isotropic, unimodal                                 & CMA-ES 30/30       & 19/9/2  & 1/10/19 & 0/20/10 \\
2  & Ellipsoidal (sep.)         & separable; unimodal; $\kappa\!\approx\!10^6$               & CMA-ES 25/30       & 27/0/3  & 4/0/26  & 5/15/10 \\
3  & Rastrigin (sep.)           & separable; regular multimodality; $\sim\!10^D$ optima      & jSO-derived 23/30          & 22/0/8  & 29/0/1  & 28/0/2  \\
4  & Bueche--Rastrigin          & asymmetric/deceptive Rastrigin; $\sim\!10^D$ optima        & jSO-derived 22/30          & 16/0/14 & 30/0/0  & 29/0/1  \\
5  & Linear Slope               & linear; optimum at domain boundary                         & 4-way tie (30/30)  & 0/30/0  & 7/23/0  & 0/30/0  \\
\midrule
6  & Attractive Sector          & strongly asymmetric unimodal                               & CMA-ES 30/30       & 29/0/1  & 9/0/21  & 0/10/20 \\
7  & Step Ellipsoidal           & plateaus; nonseparable; $\kappa\!\approx\!10^2$            & jSO-derived 30/30          & 26/0/4  & 13/0/17 & 30/0/0  \\
8  & Rosenbrock                 & long curved valley; structured dependency                  & CMA-ES 30/30       & 22/0/8  & 15/0/15 & 0/10/20 \\
9  & Rosenbrock (rot.)          & rotated curved valley; nonseparable                        & CMA-ES 30/30       & 22/0/8  & 8/0/22  & 0/8/22  \\
\midrule
10 & Ellipsoidal (rot.)         & rotated ellipsoid; $\kappa\!\approx\!10^6$                 & CMA-ES 30/30       & 28/0/2  & 7/0/23  & 0/6/24  \\
11 & Discus                     & one highly sensitive direction; $\kappa\!\approx\!10^6$    & CMA-ES 29/30       & 28/0/2  & 14/0/16 & 1/10/19 \\
12 & Bent Cigar                 & narrow smooth ridge; $\kappa\!\approx\!10^6$               & CMA-ES 27/30       & 26/0/4  & 5/0/25  & 3/3/24  \\
13 & Sharp Ridge                & nondifferentiable sharp ridge                              & CMA-ES 20/30       & 26/0/4  & 7/0/23  & 10/0/20 \\
14 & Different Powers           & nonuniform coordinate sensitivities                        & CMA-ES 25/30       & 29/0/1  & 12/0/18 & 5/1/24  \\
\midrule
15 & Rastrigin (rot.)           & nonseparable Rastrigin; $\sim\!10^D$ optima                & jSO-derived 21/30          & 22/0/8  & 14/0/16 & 21/0/9  \\
16 & Weierstrass                & rugged, repetitive multimodality                           & CMA-ES 16/30       & 16/0/14 & 0/0/30  & 4/0/26  \\
17 & Schaffers F7               & multimodal, rotated, low conditioning                      & jSO-derived 28/30          & 24/0/6  & 3/0/27  & 28/0/2  \\
18 & Schaffers F7 (ill-cond.)   & Schaffers F7 with $\kappa\!\approx\!10^3$                  & jSO-derived 30/30          & 25/0/5  & 4/0/26  & 30/0/0  \\
19 & Griewank--Rosenbrock       & highly multimodal Rosenbrock-like                          & CMA / iLS (15/30)  & 13/0/17 & 0/0/30  & 9/0/21  \\
\midrule
20 & Schwefel                   & multimodal; minima near domain corners                     & jSO-derived 18/30          & 11/0/19 & 30/0/0  & 25/0/5  \\
21 & Gallagher 101 peaks        & 101 peaks; weak global struct.; $\kappa\!\approx\!30$       & jSO-derived 19/30          & 20/1/9  & 21/1/8  & 30/0/0  \\
22 & Gallagher 21 peaks         & 21 peaks; weak global struct.; $\kappa\!\approx\!10^3$     & jSO-derived 21/30          & 23/2/5  & 14/2/14 & 27/2/1  \\
23 & Katsuura                   & highly rugged/repetitive; $>10^D$ global optima            & iL-SHADE 12/30     & 15/0/15 & 4/0/26  & 16/0/14 \\
24 & Lunacek bi-Rastrigin       & two-funnel, highly multimodal, deceptive                   & iL-SHADE 16/30     & 18/0/12 & 0/0/30  & 9/0/21  \\
\bottomrule
\end{tabularx}
\end{table}

\begin{figure*}[htpb]
    \centering
    \includegraphics[width=0.8\textwidth]{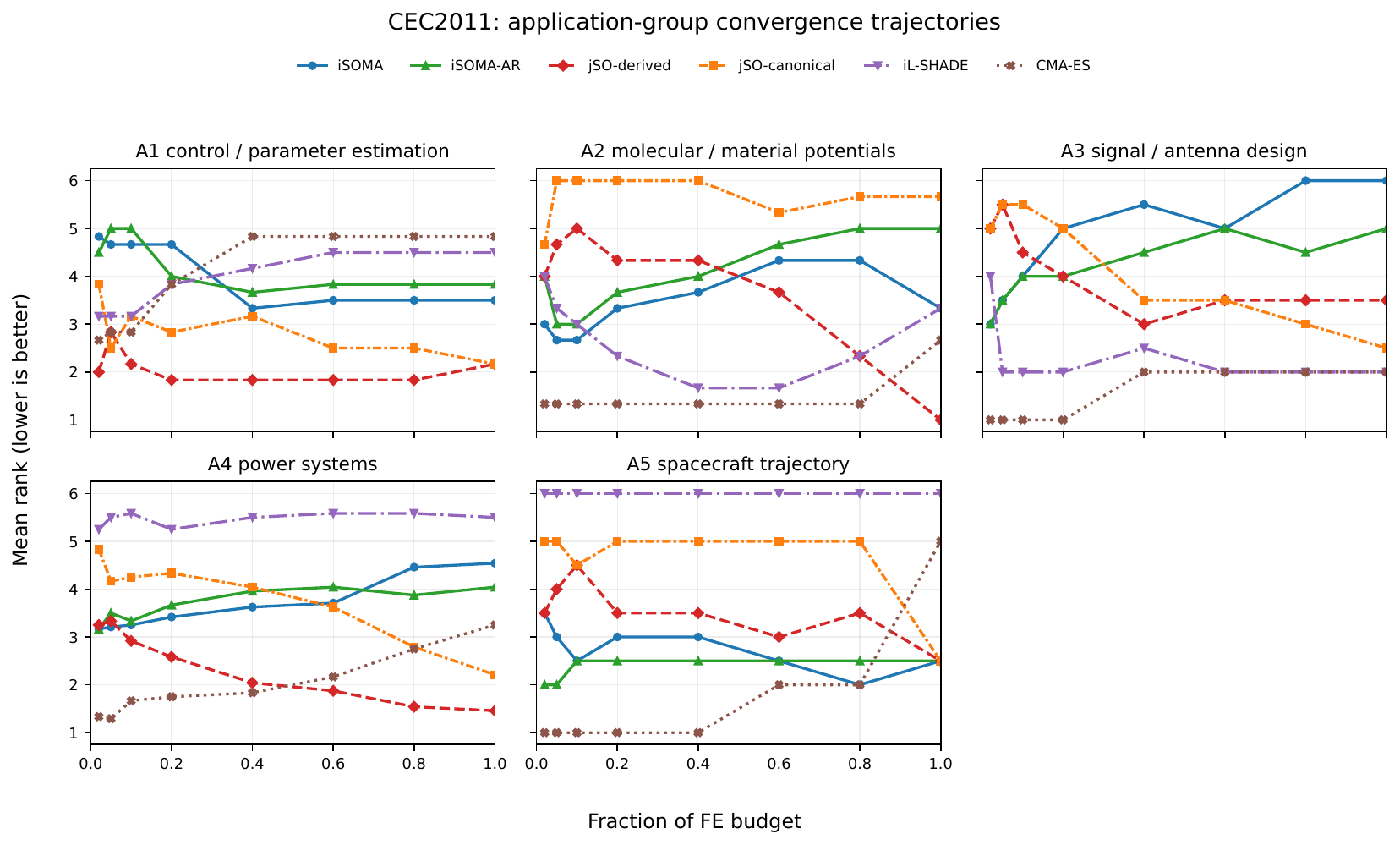}
    \caption{CEC 2011 mean-rank convergence trajectories by diagnostic application family.}
    \label{fig:cec_convergence}
\end{figure*}

\begin{figure*}[htpb]
    \centering
    \includegraphics[width=0.8\textwidth]{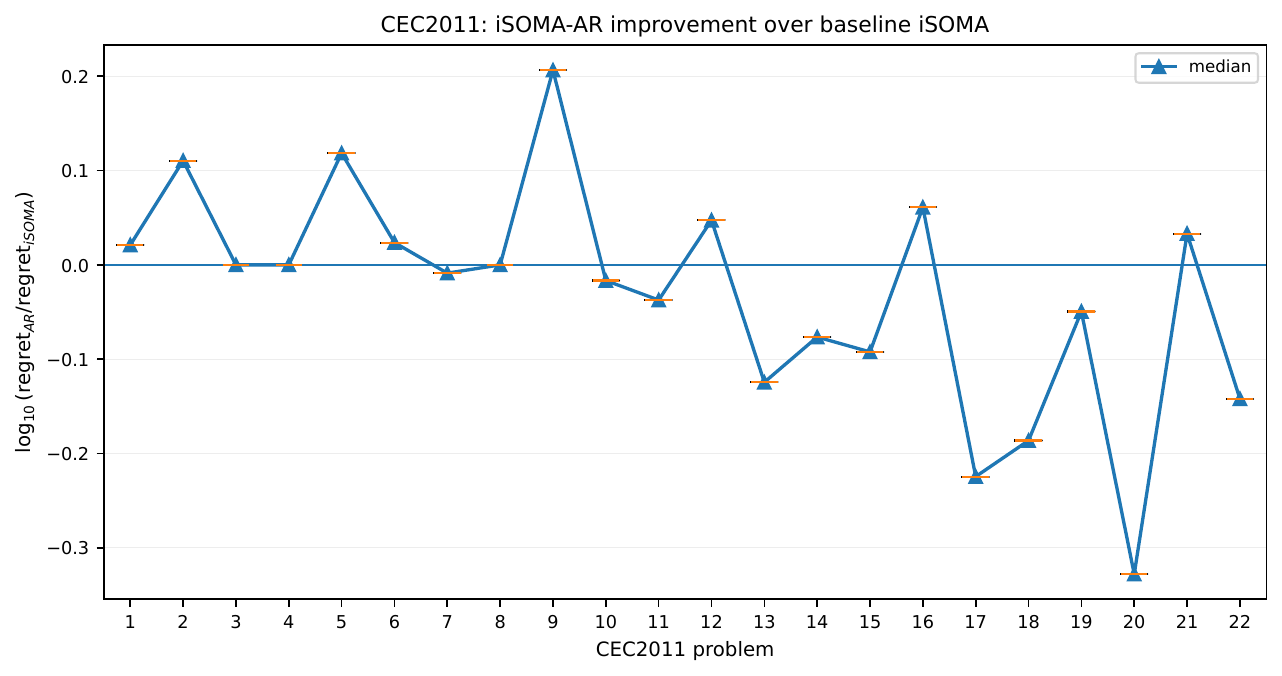}
    \caption{Problem-level $\log_{10}$ regret ratio of iSOMA-AR relative to baseline iSOMA on CEC 2011. Values below zero favor iSOMA-AR.}
    \label{fig:cec_ar_problem}
\end{figure*}

\begin{table}[htpb]
\centering
\footnotesize
\setlength{\tabcolsep}{3.5pt}
\renewcommand{\arraystretch}{0.93}
\caption{CEC 2011 problem-level outcomes at 50,000 evaluations. A dagger ($^\dagger$) marks problems for which the Minion C++ rewrite has documented discrepancies relative to the Octave/MATLAB reference.}
\label{tab:cec_problems}
\begin{tabularx}{\linewidth}{@{} r >{\raggedright\arraybackslash}X l r >{\raggedright\arraybackslash}p{2.3cm} ccc @{}}
\toprule
$F$ & Problem & Application family & $D$ & Frequent winner & AR/base & AR/iLS & jSO-derived/CMA \\
\midrule
1           & FM sound parameter estimation       & A1 Control \& param. &   6 & jSO-derived 1/1          & 0/0/1 & 1/0/0 & 1/0/0 \\
2           & Lennard--Jones potential            & A2 Materials         &  30 & jSO-derived 1/1          & 0/0/1 & 0/0/1 & 1/0/0 \\
3$^\dagger$ & Bifunctional catalyst opt. control  & A1 Control \& param. &   1 & All tied (1/1)   & 0/1/0 & 0/1/0 & 0/1/0 \\
4$^\dagger$ & Stirred-tank reactor opt. control   & A1 Control \& param. &   1 & jSO-derived 1/1          & 0/1/0 & 0/1/0 & 1/0/0 \\
5           & Tersoff potential Si(B)             & A2 Materials         &  30 & jSO-derived 1/1          & 0/0/1 & 0/0/1 & 1/0/0 \\
6           & Tersoff potential Si(C)             & A2 Materials         &  30 & jSO-derived 1/1          & 0/0/1 & 1/0/0 & 1/0/0 \\
7           & Spread-spectrum radar polyphase code & A3 Radar \& antenna  &  20 & iL-SHADE 1/1     & 1/0/0 & 0/0/1 & 0/0/1 \\
8           & Transm. network expansion planning  & A4 Power systems     &   7 & 3-way tie (1/1)  & 0/1/0 & 1/0/0 & 1/0/0 \\
9           & Large-scale transmission pricing    & A4 Power systems     & 126 & CMA-ES 1/1       & 0/0/1 & 0/0/1 & 0/0/1 \\
10          & Circular antenna array design       & A3 Radar \& antenna  &  12 & CMA-ES 1/1       & 1/0/0 & 0/0/1 & 0/0/1 \\
11          & Dynamic economic dispatch 1         & A4 Power systems     & 120 & jSO-derived 1/1          & 1/0/0 & 1/0/0 & 1/0/0 \\
12          & Dynamic economic dispatch 2         & A4 Power systems     & 240 & CMA-ES 1/1       & 0/0/1 & 1/0/0 & 0/0/1 \\
13          & Economic load dispatch 1            & A4 Power systems     &   6 & jSO-derived 1/1          & 1/0/0 & 1/0/0 & 1/0/0 \\
14          & Economic load dispatch 2            & A4 Power systems     &  13 & jSO-derived 1/1          & 1/0/0 & 0/0/1 & 1/0/0 \\
15          & Economic load dispatch 3            & A4 Power systems     &  15 & jSO-derived 1/1          & 1/0/0 & 1/0/0 & 1/0/0 \\
16          & Economic load dispatch 4            & A4 Power systems     &  40 & jSO-derived 1/1          & 0/0/1 & 1/0/0 & 1/0/0 \\
17          & Economic load dispatch 5            & A4 Power systems     & 140 & jSO-derived 1/1          & 1/0/0 & 1/0/0 & 1/0/0 \\
18          & Hydrothermal scheduling 1           & A4 Power systems     &  96 & jSO-derived 1/1          & 1/0/0 & 1/0/0 & 1/0/0 \\
19          & Hydrothermal scheduling 2           & A4 Power systems     &  96 & jSO-derived 1/1          & 1/0/0 & 1/0/0 & 1/0/0 \\
20          & Hydrothermal scheduling 3           & A4 Power systems     &  96 & jSO-derived 1/1          & 1/0/0 & 1/0/0 & 1/0/0 \\
21$^\dagger$& Messenger spacecraft trajectory     & A5 Spacecraft traj.  &  26 & jSO-derived 1/1          & 0/0/1 & 1/0/0 & 1/0/0 \\
22$^\dagger$& Cassini 2 spacecraft trajectory     & A5 Spacecraft traj.  &  22 & iSOMA-AR 1/1     & 1/0/0 & 1/0/0 & 1/0/0 \\
\bottomrule
\end{tabularx}
\end{table}

Table~\ref{tab:vqe_exact_pairwise} gives the direct exact-objective iSOMA--iSOMA-AR tests for all 16 original Q1--Q3 conditions. Table~\ref{tab:vqe_noise_ranks} summarizes condition-wise ranks under the two sampling-noise levels, while Table~\ref{tab:vqe_noise_pairwise} reports the direct noisy AR--baseline tests. The latter confirm that the strong high-noise result is a SOMA-family effect: no direct AR--baseline comparison is significant after Holm correction.

\begin{table}[htpb]
\centering
\small
\setlength{\tabcolsep}{5pt}
\renewcommand{\arraystretch}{0.92}
\caption{Direct exact-objective comparison of iSOMA-AR with baseline iSOMA. $p_{\mathrm{Holm}}$ is the Holm-adjusted two-sided Mann--Whitney $U$ result within each benchmark condition. The effect column is oriented so that positive values favor iSOMA-AR because lower error is better.}
\label{tab:vqe_exact_pairwise}
\begin{tabular}{@{}llrrrr@{}}
\toprule
Model & Condition & iSOMA median & AR median & $p_{\mathrm{Holm}}$ & Effect \\
\midrule
Q1 & $N=12$, 10k & \num{1.50e-05} & \num{1.46e-04} & 0.886 & -0.136 \\
Q1 & $N=12$, 30k & 0              & \num{1.17e-13} & 0.005 & -0.485 \\
Q1 & $N=14$, 10k & \num{6.93e-08} & \num{5.10e-07} & 0.159 & -0.341 \\
Q1 & $N=14$, 30k & 0              & 0              & 1.000 & -0.110 \\
Q1 & $N=16$, 10k & 0.458          & 0.458          & 1.000 &  0.014 \\
Q1 & $N=16$, 30k & 0.458          & 0.458          & 0.030 & -0.469 \\
\midrule
Q2 & $N=10$, 10k & 0.275          & 0.209          & 0.179          & 0.392 \\
Q2 & $N=10$, 30k & 0.206          & 0.116          & \num{3.30e-05} & 0.782 \\
Q2 & $N=12$, 10k & 0.265          & 0.201          & \num{7.01e-05} & 0.757 \\
Q2 & $N=12$, 30k & 0.231          & 0.121          & \num{2.88e-07} & 0.926 \\
\midrule
Q3 & $N=8$, 10k  & 0.767          & 0.766          & 1.000 & -0.078 \\
Q3 & $N=8$, 30k  & 0.766          & 0.766          & 0.503 &  0.216 \\
Q3 & $N=10$, 10k & 1.367          & 1.426          & 1.000 & -0.158 \\
Q3 & $N=10$, 30k & 1.367          & 0.895          & 0.219 &  0.318 \\
Q3 & $N=12$, 10k & 1.902          & 1.902          & 0.396 & -0.274 \\
Q3 & $N=12$, 30k & 1.902          & 0.999          & 0.128 &  0.358 \\
\bottomrule
\end{tabular}
\end{table}

\begin{table}[htpb]
\centering
\small
\caption{Performance metrics in the $N=12$ noisy VQE extension across low- and high-noise regimes. Lower is better for both metrics.}
\label{tab:vqe_noise_ranks}
\label{tab:vqe_selection_penalty}
\begin{tabular}{@{}lccccc@{}}
\toprule
& \multicolumn{2}{c}{(a) Condition-wise mean rank} & & \multicolumn{2}{c}{(b) Median selection penalty} \\
\cmidrule(lr){2-3} \cmidrule(lr){5-6}
Optimizer & Low noise & High noise & & Low noise & High noise \\
\midrule
iSOMA       & 3.00 & 2.17 & & 0.034 & 0.313 \\
iSOMA-AR    & 3.00 & 1.83 & & 0.046 & 0.381 \\
jSO-derived & 4.00 & 3.67 & & 0.069 & 0.620 \\
iL-SHADE    & 3.67 & 3.50 & & 0.073 & 0.583 \\
CMA-ES      & 3.83 & 4.50 & & 0.044 & 0.410 \\
SPSA        & 3.50 & 5.33 & & 0.056 & 0.707 \\
\bottomrule
\end{tabular}
\end{table}

\begin{table}[htpb]
\centering
\small
\setlength{\tabcolsep}{5pt}
\renewcommand{\arraystretch}{0.92}
\caption{Direct noisy-objective comparison of iSOMA-AR with baseline iSOMA. Endpoints are scored by their exact energy after noisy selection. No direct AR--baseline comparison remains significant after Holm correction.}
\label{tab:vqe_noise_pairwise}
\begin{tabular}{@{}llrrrr@{}}
\toprule
Model & Budget & iSOMA median & AR median & $p_{\mathrm{Holm}}$ & Effect \\
\midrule
\multicolumn{6}{@{}l}{\textit{High noise}} \\
Q1 & 10k & 1.037 & 0.958 & 1.000 &  0.181 \\
Q1 & 30k & 0.719 & 0.649 & 1.000 &  0.034 \\
Q2 & 10k & 0.593 & 0.707 & 0.898 & -0.043 \\
Q2 & 30k & 0.433 & 0.522 & 0.664 & -0.162 \\
Q3 & 10k & 2.318 & 2.425 & 0.923 &  0.018 \\
Q3 & 30k & 2.221 & 2.177 & 1.000 &  0.024 \\
\midrule
\multicolumn{6}{@{}l}{\textit{Low noise}} \\
Q1 & 10k & 0.066 & 0.087 & 1.000 &  0.008 \\
Q1 & 30k & 0.062 & 0.080 & 1.000 & -0.046 \\
Q2 & 10k & 0.317 & 0.308 & 1.000 & -0.069 \\
Q2 & 30k & 0.319 & 0.316 & 1.000 &  0.059 \\
Q3 & 10k & 1.928 & 1.927 & 1.000 & -0.062 \\
Q3 & 30k & 1.907 & 1.910 & 1.000 & -0.082 \\
\bottomrule
\end{tabular}
\end{table}

\begin{figure*}[htpb]
    \centering
    \includegraphics[width=0.75\textwidth]{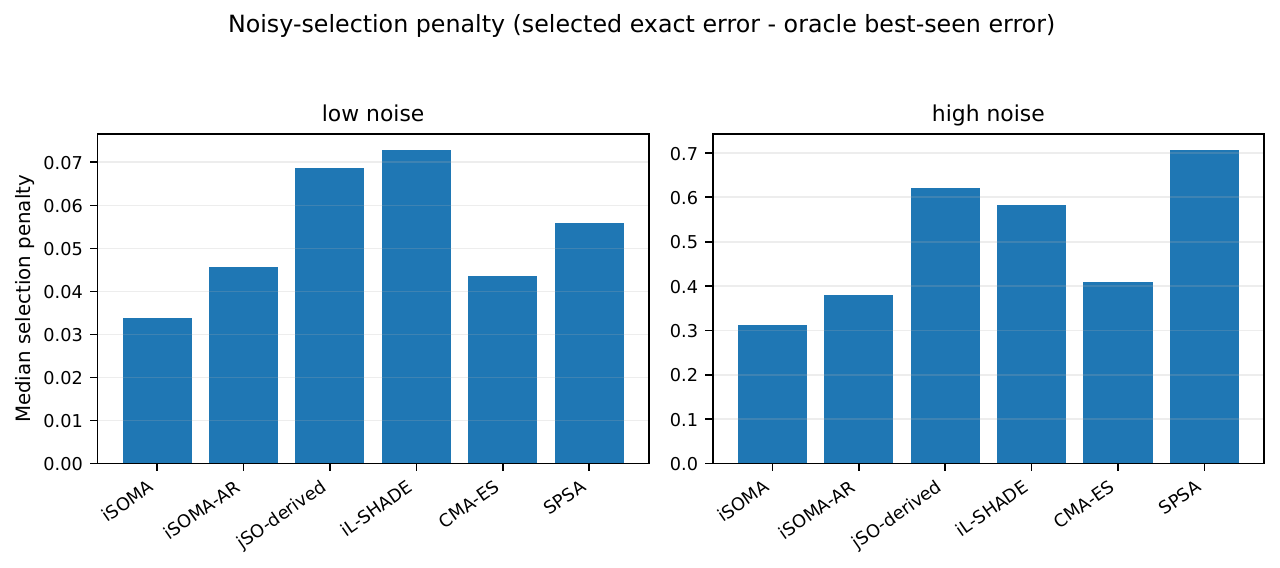}
    \caption{Noisy-selection penalty, measured as the exact error of the point selected through noisy observations minus the oracle-best exact error among all points queried in the same run. Lower values indicate that noisy selection loses less of the search progress already achieved.}
    \label{fig:vqe_selection_penalty}
\end{figure*}

\clearpage
\bibliography{references}

\end{document}